\documentclass[accepted]{uai2026} 
                        
\usepackage[american]{babel}

\usepackage{bm}
\usepackage{dsfont} 
\usepackage{bbm}
\usepackage{algorithm}
\usepackage{algpseudocode}   
\usepackage{multirow}
\usepackage{subcaption}

\newtheorem{theorem}{Theorem}

\usepackage{natbib} 
\usepackage{mathtools} 
\usepackage{booktabs} 
\usepackage{tikz} 

\title{Poisson–Gamma Modeling of Inter-Relational Dependencies in Dynamic Knowledge Graphs}

\author[1,2]{Nan~Fang}
\author[1,3]{Yijun~Wang}
\author[2]{Hao~Liao}
\author[1,4]{Sikun~Yang\thanks{Corresponding author: sikunyang@gbu.edu.cn}}

\affil[1]{%
    School of Computing and Information Technology\\
    Great Bay University\\
    Dongguan, Guangdong, China
}

\affil[2]{%
    College of Computer Science and Software Engineering\\
    Shenzhen University\\
    Shenzhen, Guangdong, China
}
\affil[3]{%
    Tsinghua University\\
    Shenzhen, Guangdong, China
}
\affil[4]{%
Guangdong Provincial Key Laboratory of Mathematical and Neural Dynamical Systems
}  
  
\begin{document}
\maketitle

\begin{abstract}

Dynamic knowledge graphs are ubiquitous in today’s AI applications, as we represent molecular structures, social relationships, and language information using these graph models. As knowledge graphs evolve over time and are often noisy and incomplete, modeling their temporal and relational dependencies becomes crucial for downstream tasks. To address these challenges, this paper proposes PGRE (Poisson–Gamma Relational Evolution), a probabilistic model for modeling inter-relational dependencies in dynamic knowledge graphs. 
PGRE represents multi-relational temporal links via a Poisson–Bernoulli formulation. It introduces Gamma-distributed latent variables to capture entity–factor associations and cross-relation dependencies mediated by shared latent communities. A Gamma Markov process further models the temporal evolution of these latent variables, enabling principled characterization of relational dynamics. 
Experiments on benchmark datasets show that PGRE achieves competitive performance in link prediction, particularly in sparse settings, while revealing meaningful relational evolution patterns in dynamic knowledge graphs.

\end{abstract}

\section{Introduction}\label{sec:intro}
Knowledge graphs (KGs)~\citep{ji2021survey} provide a fundamental representation of structured knowledge about entities and relations, and support a wide range of applications including question answering~\citep{yasunaga2021qa,chakraborty2021introduction,jia2021complex}, recommender systems~\citep{wang2019explainable,wang2021learning,wang2019kgat}, and semantic search~\citep{thingbaijam2024incorporating,xiong2017explicit,ehrlinger2016towards}.
In practical settings, however, knowledge is inherently dynamic: relations may emerge, evolve, or disappear over time.
This motivates the study of dynamic knowledge graphs (DKGs)~\citep{liang2024survey}. DKGs model multi-relational interactions as temporal event sequences to predict future or missing relations from historical observations.

Knowledge graph completion has been extensively studied over the past decade, with significant progress driven by representation learning methods based on neural architectures~\citep{schlichtkrull2018modeling,trivedi2017know,rossi2020temporal,jin2019recurrent}.
Despite their strong empirical performance, these approaches often rely on large amounts of training data and exhibit limited interpretability due to their black-box nature~\citep{chen2023tempme,seo2024self}.
Such limitations become more pronounced in sparse or small-sample settings, which are common in real-world dynamic knowledge graphs~\citep{huang2023temporal,zhou2022tgl}.

\begin{figure}[t]
    \centering
    \includegraphics[width=1\linewidth]{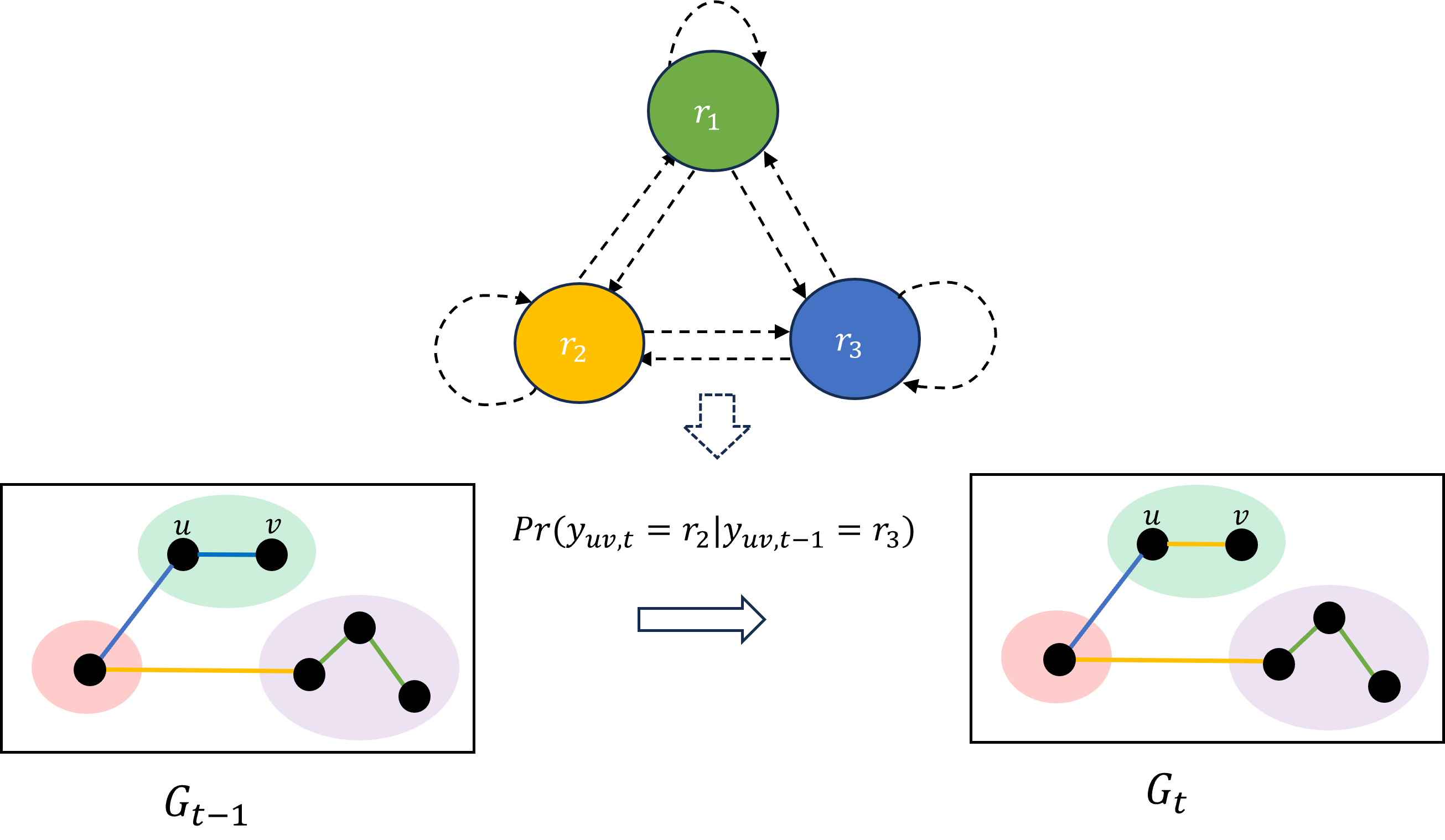}
    \caption{
    Illustration of a relation transition prior and its manifestation in dynamic knowledge graphs.
    The top panel shows an illustrative prior over relation transitions, where directed edges encode assumed transition probabilities between relation types.
    The bottom panel provides an example of how the relation between an entity pair may change from time $t\!-\!1$ to $t$, depending on its previous relation (e.g., $\Pr(y_{uv,t}=r_2 \mid y_{uv,t-1}=r_3)$).
    }
    \label{fig1:relation_prior}
\end{figure}

At the same time, relations in knowledge graphs are rarely independent.
Different relation types may exhibit structured dependencies, and relational facts observed in the past can directly influence future interactions~\citep{schlichtkrull2018modeling,trivedi2017know}.
As illustrated in Fig.~\ref{fig1:relation_prior}, such dependencies often manifest as structured relation transitions over time, where the relation of an entity pair at time $t$ depends on its state at the previous time step $t\!-\!1$.
Ignoring these inter-relational and temporal dependencies may result in an incomplete characterization of relational evolution.
Bayesian approaches provide a principled alternative by explicitly modeling uncertainty and structured dependencies~\citep{acharya2015nonparametric,yang2018poisson,pmlr-v80-yang18b,schein2019poisson,pmlr-v124-yang20a,DBLP:conf/sdm/YangZ23,DBLP:conf/aaai/YangZ24}, offering a natural framework for capturing relational dynamics while maintaining interpretability.

In this paper, we propose PGRE (Poisson--Gamma Relational Evolution), a probabilistic model for dynamic knowledge graph completion. PGRE represents multi-relational temporal links through a Poisson--Gamma latent-variable formulation and introduces a relation transition matrix within Gamma-Markov dynamics to explicitly capture inter-relational temporal dependencies. By jointly modeling entity--factor associations and relation-to-relation evolution in a unified generative framework, PGRE provides interpretable latent representations and supports tractable posterior inference for dynamic link prediction.

The main contributions of this paper are summarized as follows:
\begin{enumerate}
    \item PGRE, a probabilistic model for dynamic knowledge graph completion, is proposed to jointly model entity--factor associations and inter-relational temporal dynamics.

    \item Single-relation probabilistic modeling is extended to a multi-relational setting, enabling structured characterization of relational evolution in dynamic knowledge graphs.

    \item An efficient Gibbs sampling algorithm based on negative-binomial--logarithmic data augmentation is developed for tractable  posterior inference.

    \item Extensive experiments on benchmark datasets are conducted, demonstrating that PGRE achieves competitive or superior link prediction performance, particularly in sparse and small-sample regimes, while revealing meaningful relational evolution patterns.
\end{enumerate}

\section{Related Work}

Dynamic knowledge graph completion has been extensively studied using neural representation learning methods.
Early approaches extend static embedding models by incorporating temporal information, including TTransE~\citep{garcia2018learning}, TA-TransE and TA-DistMult~\citep{leblay2018deriving}, HyTE~\citep{dasgupta2018hyte}, DacKGR~\citep{lv2020dynamic}, and DKGE~\citep{wu2022efficiently}, which introduce time-aware embeddings or temporal constraints to model relation evolution.
Subsequent work combines graph neural networks with sequence modeling to capture both structural context and temporal dynamics.
Representative models include DyRep~\citep{trivedi2019dyrep}, RE-NET~\citep{jin2019recurrent}, and RE-GCN~\citep{li2021temporal}, which leverage attention mechanisms, recurrent units, or evolutionary aggregation to predict future relations.
More recently, large language models have been leveraged for temporal knowledge graph forecasting to improve cross-scenario generalization~\citep{bai2025g2s,tang2025anre}.
While these neural approaches achieve strong empirical performance on large datasets, they typically rely on extensive training data, offer limited interpretability and face challenges in sparse or small-sample regimes.

Probabilistic modeling provides a principled alternative for dynamic network analysis by explicitly characterizing uncertainty and temporal evolution.Early work builds on the stochastic block model (SBM)~\citep{holland1983stochastic,nowicki2001estimation} and its nonparametric extensions such as IRM~\citep{kemp2006learning} and GP-EPM~\citep{zhou2015infinite}. Dynamic extensions include the dynamic stochastic block model of \citet{matias2017statistical}, D-NGPPF~\citep{acharya2015nonparametric}, which models smoothly evolving communities via gamma-Markov chains, and DPGM~\citep{yang2018poisson}, which allows time-varying community memberships. A recent extension, G-HSEPM~\citep{yu2025tracking}, introduces hierarchical transition structures to characterize latent community evolution in dynamic networks.
However, these models typically focus on single undirected networks and do not explicitly address multi-relational dynamics.
Another line of research studies temporal interaction data by reshaping dynamic knowledge graphs into event--time count matrices, leading to models such as PGDS~\citep{schein2016poisson}, PRGDS~\citep{schein2019poisson}, and NBRGDS~\citep{huang2024negative}.
Although effective for modeling temporal intensities, these approaches capture multi-relational structure indirectly through data transformation rather than directly modeling relation interactions.
As a result, Bayesian modeling of dynamic multi-relational knowledge graphs remains relatively underexplored.

\section{The Proposed Dynamic Knowledge Graph Model}

This section begins by defining the problem setting, followed by a detailed description of the proposed Poisson–Gamma model for modeling and tracking relational evolution in dynamic knowledge graphs.

\subsection{Notation and Problem Definition}


A temporal knowledge graph (TKG) represents time-evolving relational facts as a set of time-stamped quadruples $(s,r,o,t)$, where $s$ and $o$ denote entities, $r$ denotes a relation type, and $t$ is a discrete timestamp.
The collection of events at time $t$ forms a snapshot $G_t$, and a TKG over $T$ time steps is represented as a sequence $\mathcal{G}=\{G_1,\dots,G_T\}$.


For modeling convenience, we represent the TKG as a binary tensor $\mathcal{M}\in\{0,1\}^{T\times R\times N\times N}$, where $\mathcal{M}[t][r][s][o]=1$ indicates the presence of relation $r$ from $s$ to $o$ at time $t$.
Temporal knowledge graph completion then amounts to predicting missing entries of $\mathcal{M}$.
In this work, we focus on the forecasting setting, where the objective is to infer future relations based on historical observations.
Given historical snapshots up to time $t_q-1$, the goal is to infer whether a relation $(s,r,o)$ holds at a query time $t_q$.

\subsection{Poisson–Gamma Relational Evolution}
We first describe how to represent the structure of latent communities at a single time step, and then explain how the influence of these latent communities on relation types evolves over time. Throughout the paper, we use $\mathrm{Gam}$, $\mathrm{Pois}$, $\mathrm{Dir}$, $\mathrm{Bern}$, and $\mathrm{NB}$ to denote the Gamma, Poisson, Dirichlet, Bernoulli, and negative binomial distributions, respectively.


To characterize the community structure of knowledge graphs, we associate each entity with two nonnegative \textit{entity--community factor} vectors, corresponding to its roles as a subject and as an object. These vectors capture role-specific soft affiliations with multiple latent communities, while their absolute scales are not interpreted in isolation. Instead, their relative patterns, together with the relation- and time-specific community weights, determine the strength of multi-relational interactions.
Intuitively, entities may participate in multiple types of interactions simultaneously. For example, a researcher may collaborate with some peers while advising others. Such overlapping and role-dependent patterns can be represented through different subject- and object-side affiliations with multiple latent communities.
Specifically, we assume that the dynamic knowledge graph contains $K$ latent communities. The subject-side factor loading of entity $i$ on community $k$ and the object-side factor loading of entity $j$ on community $k$ are assigned Gamma priors as $\phi_{ik} \sim \mathrm{Gam}(a_0, 1/c_i)$ and $\psi_{jk} \sim \mathrm{Gam}(a_1, 1/c_j)$, respectively, where $\mathrm{Gam}(\alpha,\theta)$ denotes a Gamma distribution with shape $\alpha$ and scale $\theta$. The hyperparameters $a_0$ and $a_1$ control the prior shapes, while $c_i$ and $c_j$ regulate the magnitudes of the subject- and object-side community factor loadings, respectively.


We assume that latent communities evolve over time and are influenced by relation types. Therefore, we introduce a time- and relation-dependent variable $\delta_{k}^{(t,r)}$ to represent the state of community $k$ under relation $r$ at time $t$. Intuitively, different relation types in a knowledge graph are not independent, and their dynamics can influence each other through shared latent community structures. Entities sharing similar semantic contexts often exhibit correlated relational behaviors over time. For example, the emergence of a “collaborates with” relation between two researchers may increase the likelihood of a subsequent “publishes with” relation within the same community. To capture such dependencies, we model the state of community $k$ under relation $r$ at time $t$ as $\delta_{k}^{(t,r)} \sim \mathrm{Gam}\left( \sum_{r'=1}^{R} \pi_{rr'} \, \delta_{k}^{(t-1,r')}, 1/\tau \right)$, which indicates that the weight of community $k$ under relation $r$ at time $t$ may be influenced by the weights of the same community across all relations at the previous time step $t-1$. The strength of this influence is determined by the coefficient vector $\Pi_r = [\pi_{r1}, \pi_{r2}, \dots, \pi_{rR}]^{T}$.
In particular, we draw the initial community weight under relation $r$ from a Gamma prior as $\delta_{k}^{(1,r)} \sim \mathrm{Gam}(\nu_r / K, 1/\tau)$. We impose a Dirichlet prior on the transition kernel as $\pi_r \sim \mathrm{Dir}(\nu_1\nu_r, \cdots, \xi\nu_r, \cdots, \nu_R\nu_r)$, and draw the hyperparameter $\nu_r$ from a Gamma prior as $\nu_r \sim \mathrm{Gam}(\gamma_0 / R, 1/\beta)$, where $\gamma_0$ is the concentration parameter and $\beta$ is the hyperparameter.

It is worth noting that as the number of latent communities $K$ approaches infinity, the hierarchical Gamma prior exhibits an inherent shrinkage property that drives the weights of redundant communities toward zero. This property enables the model to automatically infer an appropriate number of active communities from data, rather than relying on a manually fixed $K$, thereby improving flexibility and interpretability in modeling real-world dynamic knowledge graphs.

Given the subject- and object-side community factor loadings $\bm{\phi}_{i}$ and $\bm{\psi}_{j}$, together with the relation- and time-dependent community weights $\delta_{k}^{(t,r)}$, we model the probability of a link of type $r$ from subject entity $i$ to object entity $j$ at time $t$ using the Bernoulli–Poisson link function as
\begin{equation}
    \mathcal{M}[t][r][i][j] \sim \mathrm{Bern}\left(1 - \exp\left(-\sum_{k=1}^{K} \delta_{k}^{(t,r)} \phi_{ik} \psi_{jk} \right)\right) .\label{eq:bernoulli_link} 
\end{equation}
Equivalently, this formulation can be expressed as
\begin{alignat}{2}
& \mathcal{M}_{ij}^{(t,r)} &\;=\;& \mathds{1}\big(x_{ij}^{(t,r)} \geq 1\big) , \label{eq:indicator} \\
& x_{ij}^{(t,r)} &\;\sim\;& \mathrm{Pois}\Big(\sum_{k=1}^{K} \delta_{k}^{(t,r)} \phi_{ik} \psi_{jk} \Big) ,\label{eq:poisson}
\end{alignat}
where $\mathcal{M}_{ij}^{(t,r)}$, equivalent to $\mathcal{M}[t][r][i][j]$ in the tensor representation, indicates whether the event $(i, r, j)$ occurs at time $t$. This Bernoulli–Poisson formulation effectively captures the discrete and sparse nature of temporal multi-relational data and enables tractable posterior inference.

\begin{figure*}[!t]
    \centering
    \includegraphics[width=0.75\linewidth]{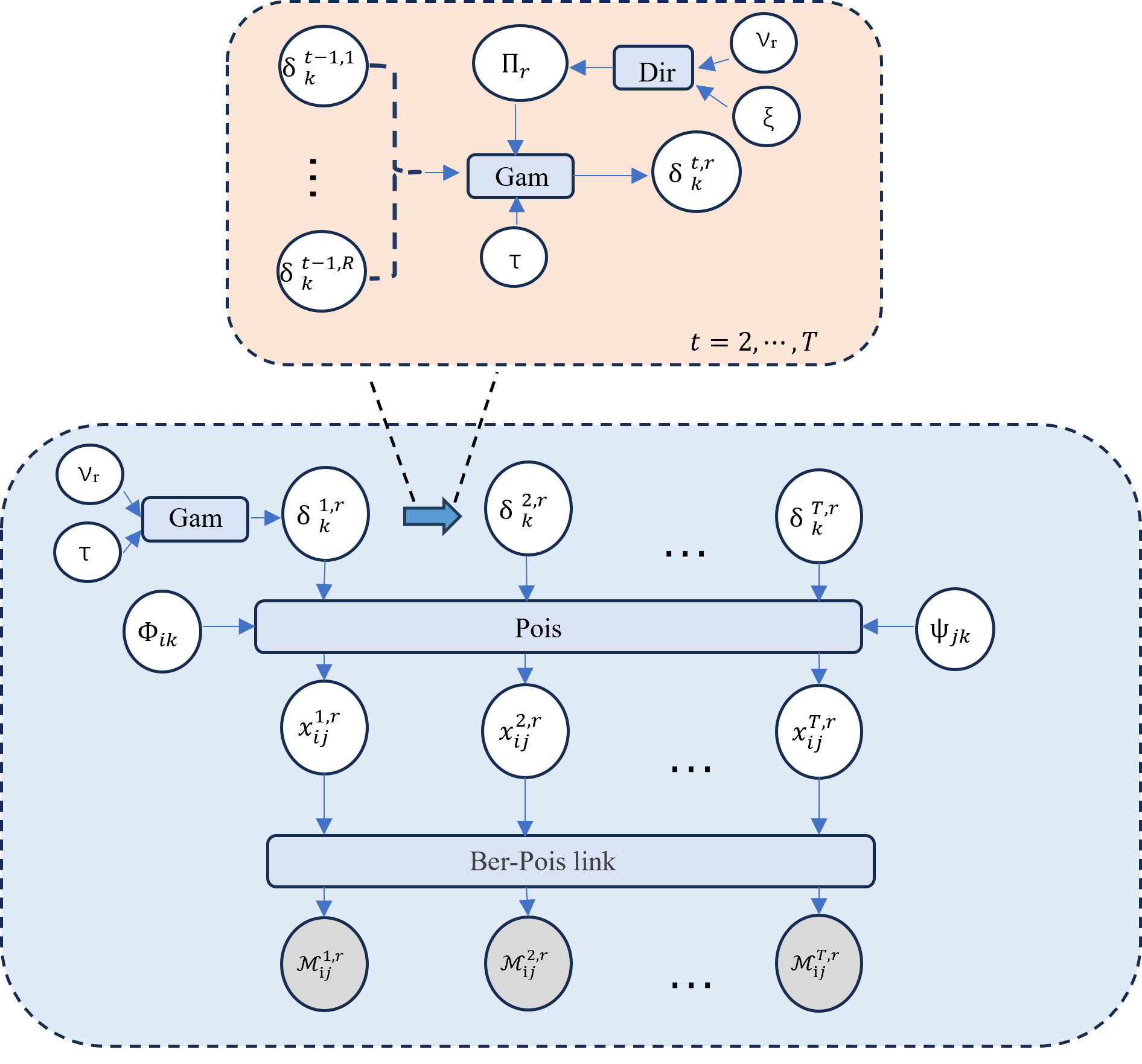}
    \caption{The overall structure of the proposed PGRE model. 
    The upper part illustrates the hierarchical Gamma–Dirichlet process governing the temporal and relational evolution of community weights $\delta_k^{(t,r)}$, while the lower part shows the Poisson–Bernoulli generative process for multi-relational links over time.}
    \label{Working_Diagram}
\end{figure*}

The full generative Poisson-Gamma relational model is specified as follows.
\begin{alignat}{2}
& \mathcal{M}_{ij}^{(t,r)} &\;=\;& \mathbbm{1}\big(x_{ij}^{(t,r)} \geq 1\big), \\
& x_{ij}^{(t,r)} &\;\sim\;& \mathrm{Pois}\Big(\sum_{k=1}^{K} \delta_{k}^{(t,r)} \phi_{ik} \psi_{jk} \Big), \\
& \phi_{ik} &\;\sim\;& \mathrm{Gam}(a_0, 1/c_i), \quad
  \psi_{jk} \;\sim\; \mathrm{Gam}(a_1, 1/c_j), \\
& \delta_{k}^{(t,r)} &\;\sim\;& \mathrm{Gam}\Big( \sum_{r'=1}^{R} \pi_{rr'} \, \delta_{k}^{(t-1,r')}, 1/\tau \Big), \quad t = 2,\dots,T, \\
& \delta_{k}^{(1,r)} &\;\sim\;& \mathrm{Gam}(\nu_r / K, 1/\tau), \\
& \pi_r &\;\sim\;& \mathrm{Dir}(\nu_1\nu_r, \dots, \xi\nu_r, \dots, \nu_R\nu_r), \\
& \nu_r &\;\sim\;& \mathrm{Gam}(\gamma_0 / R, 1/\beta).
\end{alignat}

Here, $c_i$, $c_j$, $\xi$, and $\beta$ are assigned Gamma priors, while hyperparameters such as $a_0$, $a_1$, and $\gamma_0$ are themselves drawn from Gamma distributions.
The overall structure of the proposed PGRE model is illustrated in Fig.~\ref{Working_Diagram}.

\section{Inference}
\label{Inference}
This section outlines the Gibbs sampling procedure for PGRE by presenting the posterior updates of key model parameters, with full derivations deferred to the supplementary material. The following standard results are recalled for completeness and are
not claimed as new theoretical contributions~\citep{zhou2015infinite}.

\begin{theorem}
\label{thm:gp}
Let $\lambda \sim \mathrm{Gam}\!\left(r, \frac{p}{1-p}\right)$ and $y \mid \lambda \sim \mathrm{Pois}(\lambda)$.
Then the marginal distribution of $y$ is $\mathrm{NB}(r,p)$.
Equivalently, a negative binomial random variable admits a gamma--Poisson mixture representation.
\end{theorem}

\begin{theorem}
\label{thm:pl}
The Poisson--logarithmic bivariate representation links the negative binomial distribution to the Poisson--CRT construction.
If $y \sim \mathrm{NB}(r,p)$ and $l \sim \mathrm{CRT}(y,r)$, then $y=\sum_{s=1}^l u_s$ with $u_s \sim \mathrm{Logarithmic}(p)$ and $l \sim \mathrm{Pois}\!\left(-r\log(1-p)\right)$.

\end{theorem}

\begin{theorem}
\label{thm:pm}
Let $y = \sum_{n=1}^N y_n$, where $y_n \sim \mathrm{Pois}(\lambda_n)$ are independent Poisson random variables with rates $\lambda_n$. Then, according to the Poisson–multinomial equivalence, $(y_1, \dots, y_N)\mid y \sim \mathrm{Mult}\left(y; \frac{\lambda_1}{\sum_n \lambda_n}, \dots, \frac{\lambda_N}{\sum_n \lambda_n}\right)$ and $y \sim \mathrm{Pois}\left(\sum_{n=1}^N \lambda_n \right)$.
\end{theorem}

\noindent \textbf{Sampling latent counts $x_{ijk}^{(t,r)}$.}
By Theorem~\ref{thm:pm}, the latent counts are allocated across communities as
\begin{equation}
(x_{ijk}^{(t,r)} \mid -) \sim \mathrm{Mult}\!\left(
x_{ij}^{(t,r)},
\frac{\delta_{k}^{(t,r)} \phi_{ik} \psi_{jk}}
{\sum_{k'=1}^{K}\delta_{k'}^{(t,r)} \phi_{ik'} \psi_{jk'}}
\right).
\end{equation}

\noindent \textbf{Sampling community factor loadings $\phi_{ik}$ and $\psi_{jk}$.}
By Gamma--Poisson conjugacy, both subject- and object-side factor loadings admit closed-form Gamma updates:

{\footnotesize
\begin{equation}
(\phi_{ik}\mid -)\sim
\mathrm{Gam}\!\left(
a_0+\sum_{t,r,j}x_{ijk}^{(t,r)},
\left(c_i+\sum_{t,r,j}\delta_k^{(t,r)}\psi_{jk}\right)^{-1}
\right),
\end{equation}

\begin{equation}
(\psi_{jk}\mid -)\sim
\mathrm{Gam}\!\left(
a_1+\sum_{t,r,i}x_{ijk}^{(t,r)},
\left(c_j+\sum_{t,r,i}\delta_k^{(t,r)}\phi_{ik}\right)^{-1}
\right).
\end{equation}
}

\noindent \textbf{Sampling community weights $\delta_{k}^{(t,r)}$.} 
For the final time step $t = T$, applying the Gamma--Poisson conjugacy yields the following posterior update for $\delta_{k}^{(T,r)}$.

{\footnotesize
\begin{equation}
(\delta_k^{(T,r)}\mid-)\sim
\mathrm{Gam}\!\left(
x_{\cdot\cdot k}^{(T,r)}
+\sum_{r'=1}^{R}\pi_{rr'}\delta_k^{(T-1,r')},
(\tau+s_k)^{-1}
\right).
\end{equation}
}

To incorporate forward information, we marginalize 
$\delta_k^{(t+1,r)}$ via Theorem~\ref{thm:gp} and introduce
auxiliary variables
\begin{equation}
    l_{k}^{(t,\cdot r)} \sim \text{CRT} \left( x_{\cdot \cdot k}^{(t,r)} + l_{k}^{(t+1,\cdot r)}, \sum_{r'=1}^{R} \pi_{rr'} \delta_{k}^{(t-1,r')} \right),
\end{equation}
which propagate forward information across time.

Combining backward counts and forward augmentation,
the conditional posterior of $\delta_k^{(t,r)}$ is
\begin{equation}
\begin{aligned}
    (\delta_k^{(t,r)}|-) &\sim \text{Gam} \left( x_{\cdot \cdot k}^{(t,r)} + l_{k}^{(t+1,\cdot r)} + \sum_{r'}^{R}\pi_{rr'}\delta_{k}^{(t-1,r')}, \right. \\
    &\quad \left. 1 / (\tau + s_k - \ln(1 - \rho_k^{(t+1)})) \right),
\end{aligned}
\end{equation}

The auxiliary parameter $\rho_k^{(t)}$ follows the recursion
\begin{equation}
\rho_k^{(t)} =
\frac{s_k - \ln(1-\rho_k^{(t+1)})}
{\tau + s_k - \ln(1-\rho_k^{(t+1)})}.
\end{equation}

\noindent \textbf{Sampling the transition matrix $\pi_r$.}  
After marginalizing out $\delta$, the auxiliary counts follow a multinomial distribution $(l_{k}^{(t,1r)}, \dots, l_{k}^{(t,Rr)}) \sim \text{Mult}\left(l_{k}^{(t,\cdot r)}, (\pi_{1r}, \dots, \pi_{Rr})\right)$. By the Dirichlet–multinomial conjugacy, the posterior distribution of $\pi_r$ is given by

\begin{equation}
\begin{aligned}
    (\pi_r|-) \sim \text{Dir}\Bigl(
        & \nu_1 \nu_r + \sum_{t=2}^{T} \sum_{k=1}^{K} l_{k}^{(t,1r)}, \ \dots, \\
        & \xi \nu_r + \sum_{t=2}^{T} \sum_{k=1}^{K} l_{k}^{(t,rr)}, \ \dots, \\
        & \nu_R \nu_r + \sum_{t=2}^{T} \sum_{k=1}^{K} l_{k}^{(t,Rr)}
    \Bigr).
\end{aligned}
\end{equation}

\noindent \textbf{Sampling $\nu_r$ and $\xi$.}  
We marginalize over $\Pi$ to obtain a Dirichlet–multinomial distribution

\begin{equation}
\begin{aligned}
    & ( l_{\cdot}^{(\cdot,1r)}, \dots, l_{\cdot}^{(\cdot,Rr)} ) \\
    &\quad \sim \text{DirMult}\Bigl(
        l_{\cdot}^{(\cdot,\cdot r)}, \ 
        ( \nu_1\nu_r, \dots, \xi\nu_r, \dots, \nu_R\nu_r )
    \Bigr),
\end{aligned}
\end{equation}
where $l_{\cdot}^{(\cdot,\cdot r)} = \sum_{t=1}^{T} \sum_{k=1}^{K} \sum_{r_1=1}^{K}l_{k}^{(t,r_1 r)}$. 

By introducing a beta-distributed auxiliary variable $q_r$, the Dirichlet–multinomial distribution can be rewritten as a negative binomial distribution:
\begin{equation}
    q_r \sim \text{Beta}\left( l_{\cdot}^{(\cdot,\cdot r)}, \, \nu_r \left( \xi + \sum_{r' \neq r} \nu_{r'} \right) \right).
\end{equation}

Then, $l_{\cdot}^{(\cdot,rr)}$ and $l_{\cdot}^{(\cdot, rr')}$ follow negative binomial distributions as 
$l_{\cdot}^{(\cdot,rr)} \sim \mathrm{NB}(\xi\nu_r, q_r)$ and 
$l_{\cdot}^{(\cdot, rr')} \sim \mathrm{NB}(\nu_{r}\nu_{r'}, q_r)$.
To further facilitate posterior inference, we introduce auxiliary count variables $h_{rr}$ and $h_{rr'}$ via the Chinese Restaurant Table (CRT) distribution:
\begin{equation}
\begin{aligned}
    h_{rr} &\sim \text{CRT}\left( l_{\cdot}^{(\cdot,rr)}, \, \xi \nu_r \right), \\
    h_{rr'} &\sim \text{CRT}\left( l_{\cdot}^{(\cdot,rr')}, \, \nu_{r}\nu_{r'} \right).
\end{aligned}
\end{equation}

Using the gamma–Poisson conjugacy, $\xi$ is then sampled as
\begin{equation}
    (\xi|-) \sim \mathrm{Gam}\left( b_0 + \sum_{r=1}^{R} h_{rr}, \, \frac{1}{ e_0 - \sum_{r=1}^{R} \nu_r \ln (1 - q_r) } \right).
\end{equation}

\begin{table*}[htbp]
\centering
\caption{Performance comparison on three dynamic knowledge graph datasets,
evaluated using AUC-PR and AUC-ROC.
Best results for each metric are shown in \textbf{bold}.}
\setlength{\tabcolsep}{3pt}
\begin{tabular}{lcccccc}
\toprule
\multirow{2}{*}{Model} & \multicolumn{2}{c}{ICEWS18} & \multicolumn{2}{c}{GDELT} & \multicolumn{2}{c}{WIKI} \\
\cmidrule(lr){2-3}\cmidrule(lr){4-5}\cmidrule(lr){6-7}
 & AUC-PR & AUC-ROC & AUC-PR & AUC-ROC & AUC-PR & AUC-ROC \\
\midrule
PGDS        & $0.082 \pm 0.004$ & $0.863 \pm 0.001$ & $0.124 \pm 0.014$ & $0.836 \pm 0.008$ & $0.126 \pm 0.006$ & $0.854 \pm 0.004$ \\
PRGDS       & $0.079 \pm 0.003$ & $0.865 \pm 0.002$ & $0.136 \pm 0.016$ & $0.843 \pm 0.010$ & $0.134 \pm 0.006$ & $0.889 \pm 0.002$ \\
NBRGDS      & $0.071 \pm 0.005$ & $0.876 \pm 0.004$ & $0.102 \pm 0.019$ & $0.851 \pm 0.009$ & $0.437 \pm 0.007$ & $0.949 \pm 0.003$ \\
\midrule
Know-Evolve & $0.072 \pm 0.005$ & $0.907 \pm 0.002$ & $0.105 \pm 0.017$ & $0.805 \pm 0.012$ & $0.554 \pm 0.014$ & $0.976 \pm 0.007$ \\
DyRep       & $0.075 \pm 0.006$ & $0.914 \pm 0.003$ & $0.114 \pm 0.012$ & $0.824 \pm 0.011$ & $0.598 \pm 0.015$ & $0.984 \pm 0.008$ \\
RENet       & $0.080 \pm 0.005$ & $0.916 \pm 0.004$ & $0.108 \pm 0.019$ & $0.837 \pm 0.011$ & $\mathbf{0.727 \pm 0.005}$ & $\mathbf{0.998 \pm 0.001}$ \\
\midrule
D-NGPPF    & $0.127 \pm 0.004$ & $0.930 \pm 0.004$ & $0.121 \pm 0.012$ & $0.880 \pm 0.011$ & $0.582 \pm 0.003$ & $0.987 \pm 0.003$ \\
DPGM        & $0.105 \pm 0.004$ & $0.922 \pm 0.003$ & $0.108 \pm 0.015$ & $0.874 \pm 0.008$ & $0.498 \pm 0.006$ & $0.989 \pm 0.002$ \\
G-HSEPM    & $0.120 \pm 0.002$ & $0.930 \pm 0.002$ & $0.119 \pm 0.016$ & $0.878 \pm 0.005$ & $0.556 \pm 0.006$ & $0.987 \pm 0.004$ \\
\midrule
PGRE 
& $\mathbf{0.198 \pm 0.004}$
& $\mathbf{0.936 \pm 0.003}$
& $\mathbf{0.277 \pm 0.015}$
& $\mathbf{0.902 \pm 0.004}$
& $0.607 \pm 0.003$ 
& $0.998 \pm 0.002$ \\
\bottomrule
\end{tabular}
\label{tab:relation-prediction}
\end{table*}

Next, we define 
\[
    n_r = h_{rr} + \sum_{r_1 \neq r}h_{r_{1}r} + \sum_{r_2 \neq r}h_{rr_{2}} + \sum_{k=1}^{K}l_{k}^{(1,r \cdot)},
\]
where $l_{k}^{(1,r)} \sim \text{Pois} (-\tau\nu_r\ln{(1-\rho_k^{(1)})})$. Using the Poisson additive property and gamma–Poisson conjugacy, we have

\begin{equation}
    (\nu_r|-) \sim \text{Gam} (\frac{\gamma_0}{R} + n_r, 1/(\beta + t_r)),
\end{equation}
where $t_r = -\ln(1 - q_r)\left(\xi + \sum_{r_{1} \neq r} \nu_{r_{1}}\right) 
- \sum_{r_{2} \neq r} \ln(1 - q_{r_{2}}) \nu_{r_{2}} 
- \sum_{k=1}^{K} \frac{1}{K} \ln(1 - \rho_k^{(1)})$.

\section{Experiments}

This section evaluates PGRE through comparative performance analysis, ablation studies, and parameter visualizations to assess both effectiveness and interpretability.

\subsection{Relation Prediction}

\paragraph{Datasets.}
We evaluated PGRE on three widely used temporal knowledge graph datasets: ICEWS18~\citep{boschee2015icews}, GDELT~\citep{leetaru2013gdelt}, and WIKI~\citep{leblay2018deriving}. To construct moderate-scale subgraphs with sufficient relational--temporal support, we first ranked relation types according to their total numbers of observed events and retained the most frequent relations, resulting in 10 relations for ICEWS18, 8 for GDELT, and 15 for WIKI. Within each relation-filtered subgraph, entity activity was measured by the total number of events in which an entity appeared as either a subject or an object. We then retained the 500 most active entities and removed events whose subject or object was not included in the selected entity set. This preprocessing strategy keeps Gibbs inference computationally manageable while ensuring sufficient observations for estimating inter-relational temporal dependencies.

\paragraph{Baselines.}
We compared PGRE with representative methods from three categories:
probabilistic models for dynamic networks (DPGM~\citep{yang2018poisson}, D-NGPPF~\citep{acharya2015nonparametric}, G-HSEPM~\citep{yu2025tracking}),
probabilistic models for dynamic count data (PGDS~\citep{schein2016poisson}, PRGDS~\citep{schein2019poisson}, NBRGDS~\citep{huang2024negative}),
and neural models for temporal knowledge graphs (Know-Evolve~\citep{trivedi2017know}, DyRep~\citep{trivedi2019dyrep}, RENet~\citep{jin2019recurrent}).

\paragraph{Experimental Settings.}
Each dataset was divided into three temporal segments:
$\text{train} \ (t = 1, \dots, T-2)$,
$\text{valid} \ (t = T-1)$,
and $\text{test} \ (t = T)$,
where the prediction query time is $t_q = T$.
Given historical events up to time $t_q-1$, models predict the probabilities of events occurring at time $t_q$.
Performance is evaluated using AUC-ROC and AUC-PR. More experimental details are provided in the supplementary material.

\begin{figure*}[t]
    \centering

    \begin{subfigure}[t]{0.38\textwidth}
        \centering
        \includegraphics[width=\linewidth]{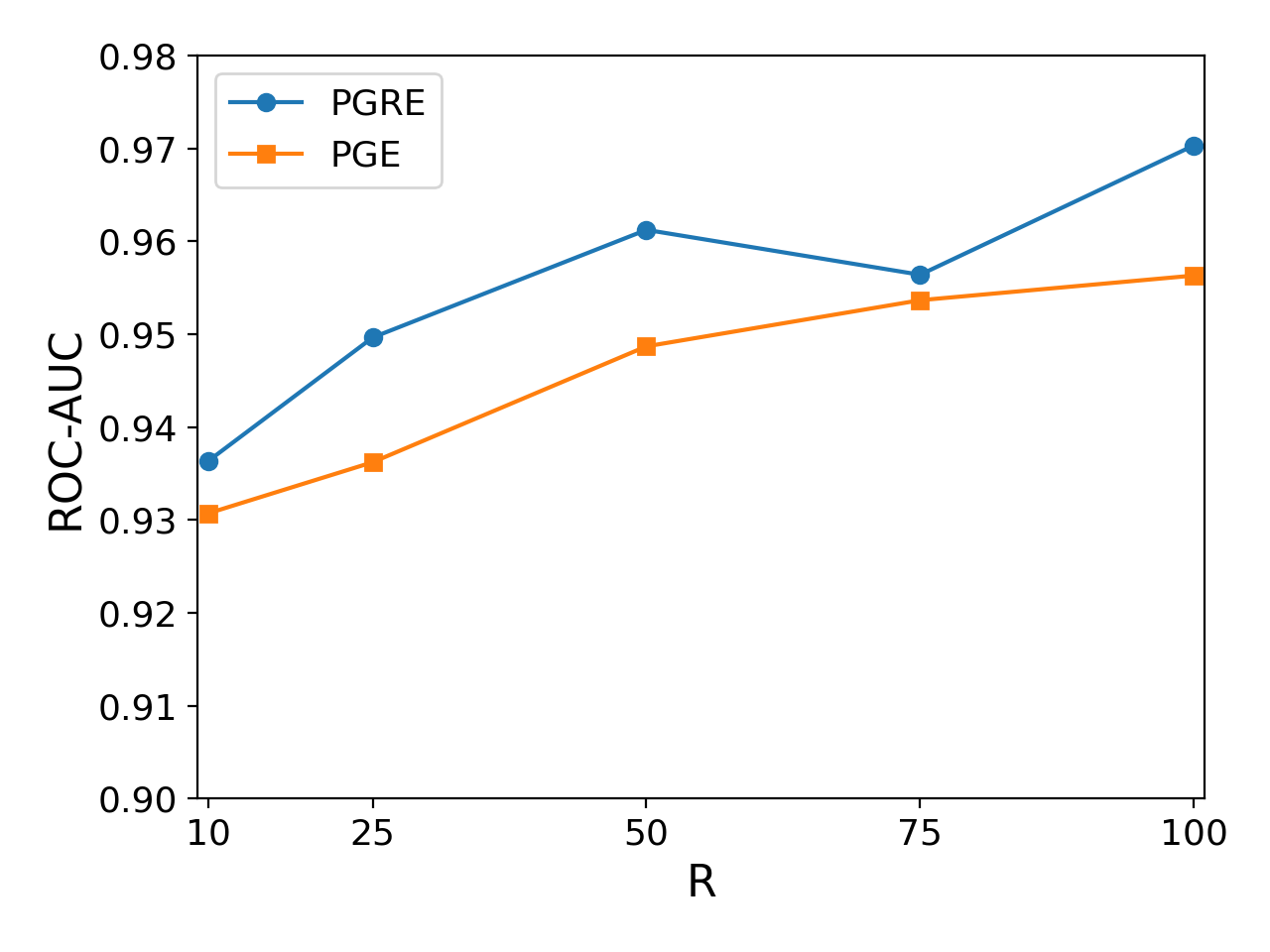}
        \subcaption{AUC-ROC: PGRE vs. PGE}
        \label{fig:roc_pgre_pge}
    \end{subfigure}
    \hspace{1em}
    \begin{subfigure}[t]{0.38\textwidth}
        \centering
        \includegraphics[width=\linewidth]{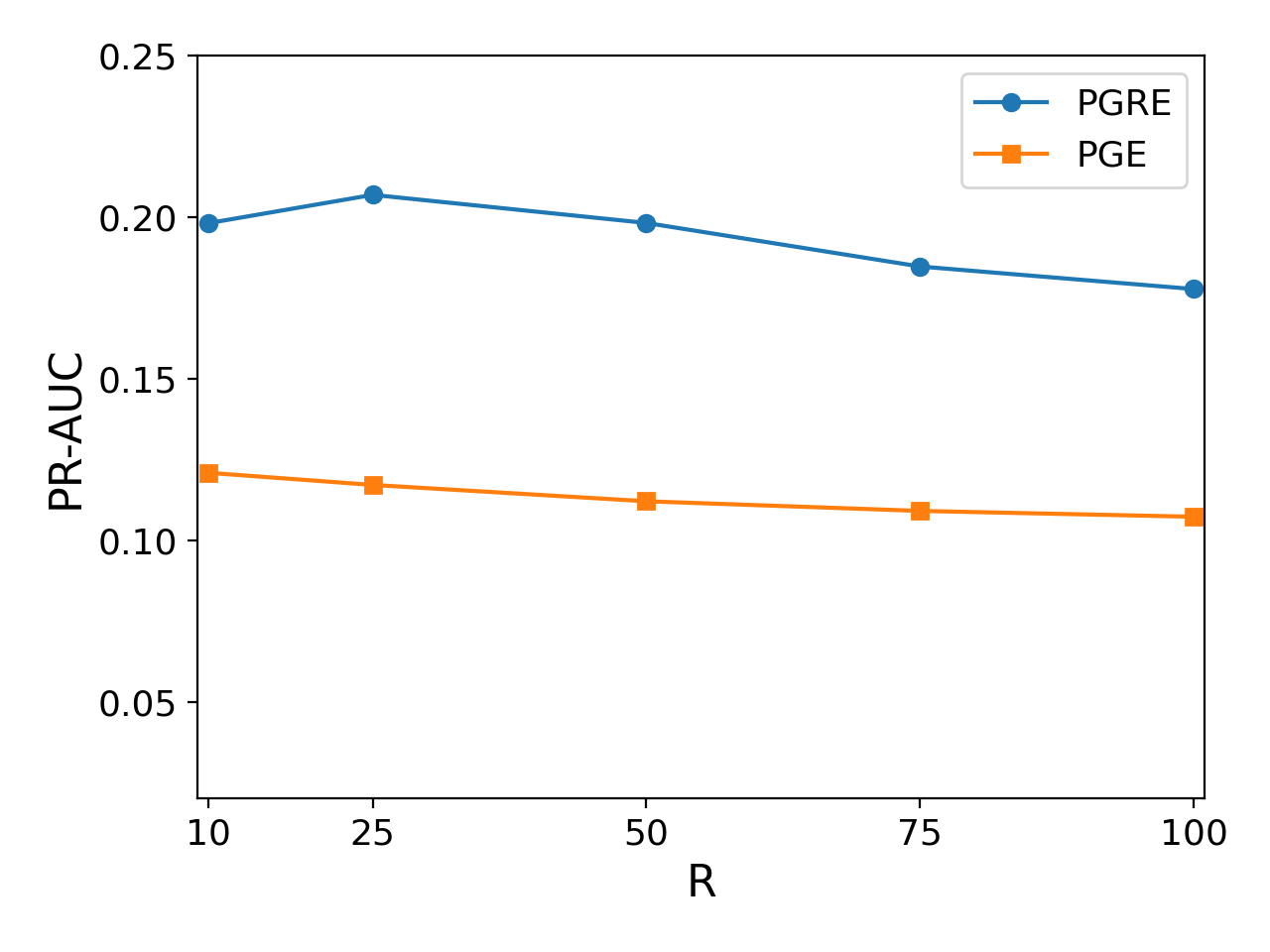}
        \subcaption{AUC-PR: PGRE vs. PGE}
        \label{fig:pr_pgre_pge}
    \end{subfigure}

    \medskip

    \begin{subfigure}[t]{0.38\textwidth}
        \centering
        \includegraphics[width=\linewidth]{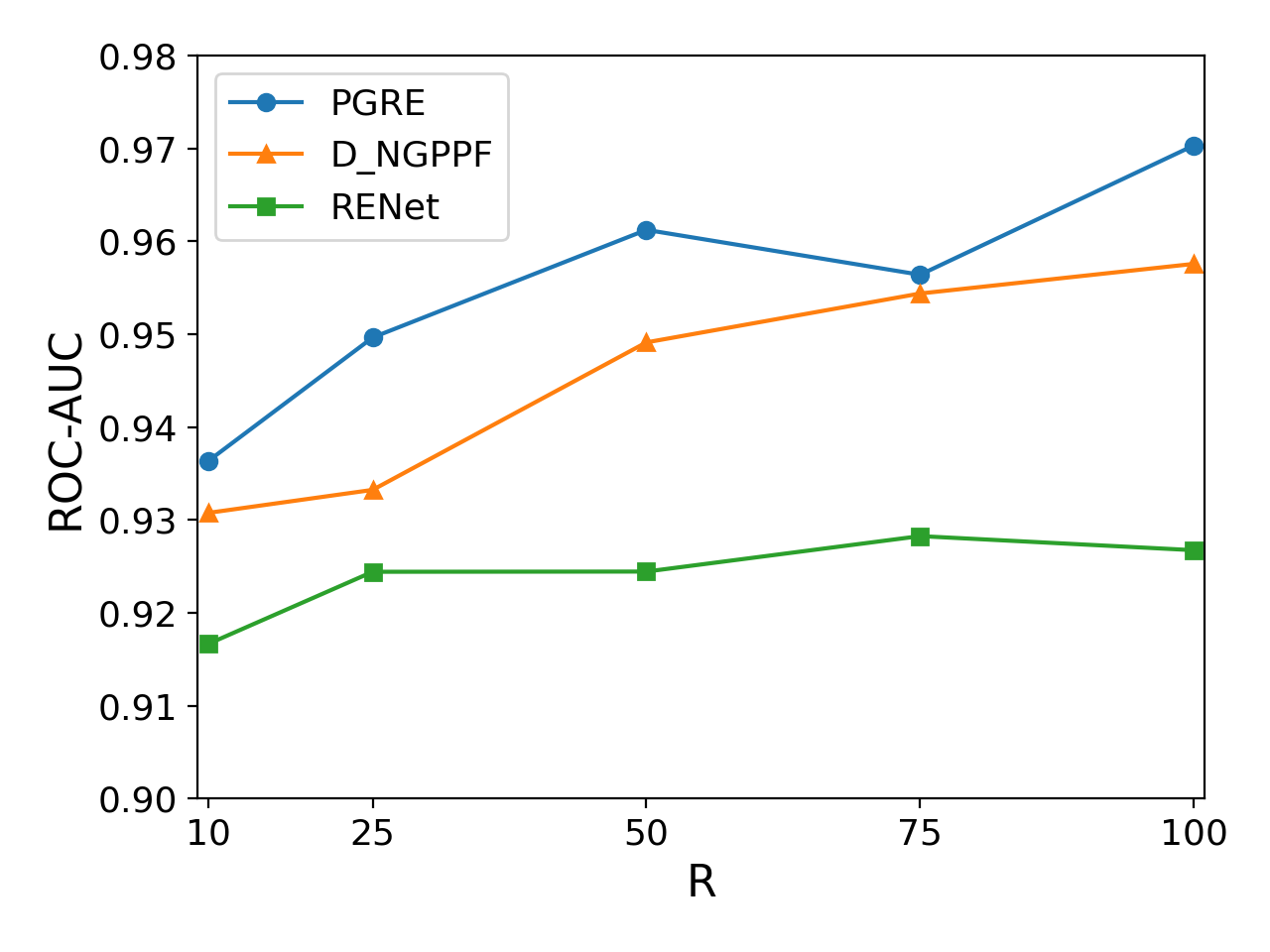}
        \subcaption{AUC-ROC: PGRE vs. D-NGPPF and RENet}
        \label{fig:roc_pgre_dngppf_renet}
    \end{subfigure}
    \hspace{1em}
    \begin{subfigure}[t]{0.38\textwidth}
        \centering
        \includegraphics[width=\linewidth]{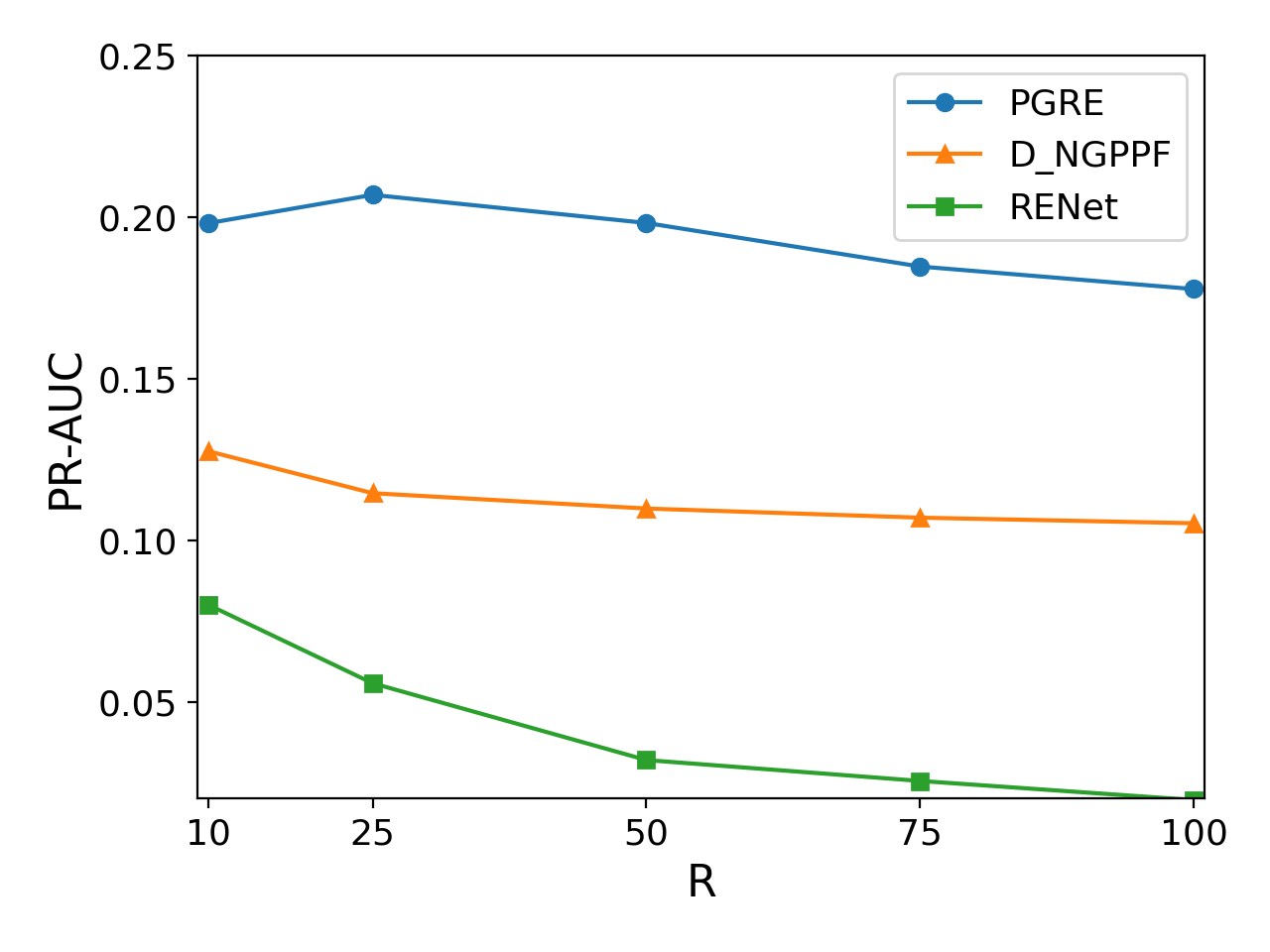}
        \subcaption{AUC-PR: PGRE vs. D-NGPPF and RENet}
        \label{fig:pr_pgre_dngppf_renet}
    \end{subfigure}

    \caption{
    Comparison of AUC-ROC and AUC-PR under different relation counts 
    $R \in \{10, 25, 50, 75, 100\}$. 
    The first row compares PGRE with its variant PGE (without the relation transition matrix), 
    while the second row compares PGRE with two strong baselines, D-NGPPF and RENet.
    }
    \label{fig:auc_comparison}
\end{figure*}

\paragraph{Results on Relation Prediction.}
Table~\ref{tab:relation-prediction} summarizes the relation prediction performance on ICEWS18, GDELT, and WIKI. 
PGRE consistently achieves competitive results across all datasets, with particularly large improvements in AUC-PR on ICEWS18 and GDELT, both of which exhibit sparse and highly multi-relational structures. 
Even on WIKI, where relational dependencies are weaker, PGRE attains the second-best AUC-PR, indicating robust generalization across varying relational densities.

The superior AUC-PR performance of PGRE primarily stems from two modeling choices. 
First, PGRE captures latent transition dependencies among relations through the transition matrix $\Pi_{rr'}$, enabling the model to represent how different relations evolve and interact over time. 
Second, the explicit modeling of directed head–tail entity roles allows PGRE to capture asymmetric temporal dynamics that are ignored by conventional single-relation models.

Compared with both count-based probabilistic methods and neural temporal models, PGRE is more robust under sparse multi-relational settings. 
While neural models often achieve higher AUC-ROC by focusing on frequent patterns, PGRE maintains substantially better AUC-PR by leveraging structured Bayesian priors that couple relations and entities, resulting in improved recall and interpretability.

\begin{table}[t]
\centering
\caption{Average time per iteration (seconds) on different datasets.}
\label{tab:runtime}
\begin{tabular}{l c c c}
\toprule
Model & ICEWS18 & GDELT & WIKI \\
\midrule
PGDS     
 & 73.28 
 & 135.52 
 & 51.91 \\
PRGDS    
 & 75.07 
 & 117.54
 & 54.85 \\
NBRGDS   
 & 72.11
 & \textbf{79.06} 
 & 46.55 \\
D-NGPPF 
 & 140.70 
 & 329.81 
 & 81.50 \\
DPGM     
 & 172.78 
 & 459.88 
 & 88.86 \\
G-HSEPM 
 & 204.01 
 & 339.26 
 & 145.98 \\
PGRE     
 & \textbf{71.70} 
 & 185.16 
 & \textbf{16.19} \\
\bottomrule
\end{tabular}
\end{table}

Table~\ref{tab:runtime} reports the average time per iteration on three benchmarks.
PGRE achieves the lowest runtime on the sparsest dataset (WIKI) and remains competitive on denser datasets such as ICEWS18 and GDELT.
Overall, PGRE provides a favorable trade-off between computational efficiency and modeling expressiveness in multi-relational temporal settings.

\subsection{Ablation and Interpretability Analysis}
\begin{figure*}[t]
    \centering
    \includegraphics[width=0.9\linewidth]{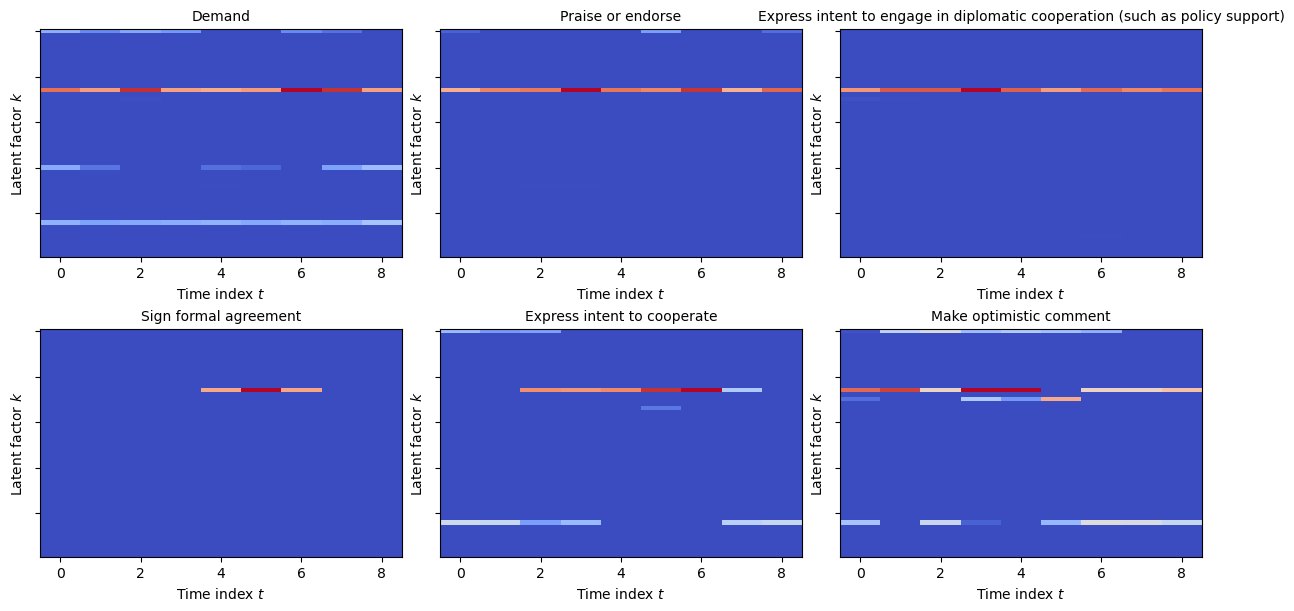}
    \caption{Temporal evolution of latent factors for six representative relations: 
    \textit{Demand}, \textit{Praise or endorse}, \textit{Express intent to engage in diplomatic cooperation}, 
    \textit{Sign formal agreement}, \textit{Express intent to cooperate}, and 
    \textit{Make optimistic comment}, $\delta_k^{t,r}/ (\sum_{k=1}^K\delta_k^{t,r}) $. 
    The $y$-axis represents latent factor indices (150--199) and the $x$-axis represents time steps $t$. 
    Redder colors indicate higher activity levels.}
    \label{fig:rk}
\end{figure*}

\begin{figure}[t]
    \centering
    \includegraphics[width=0.6\linewidth]{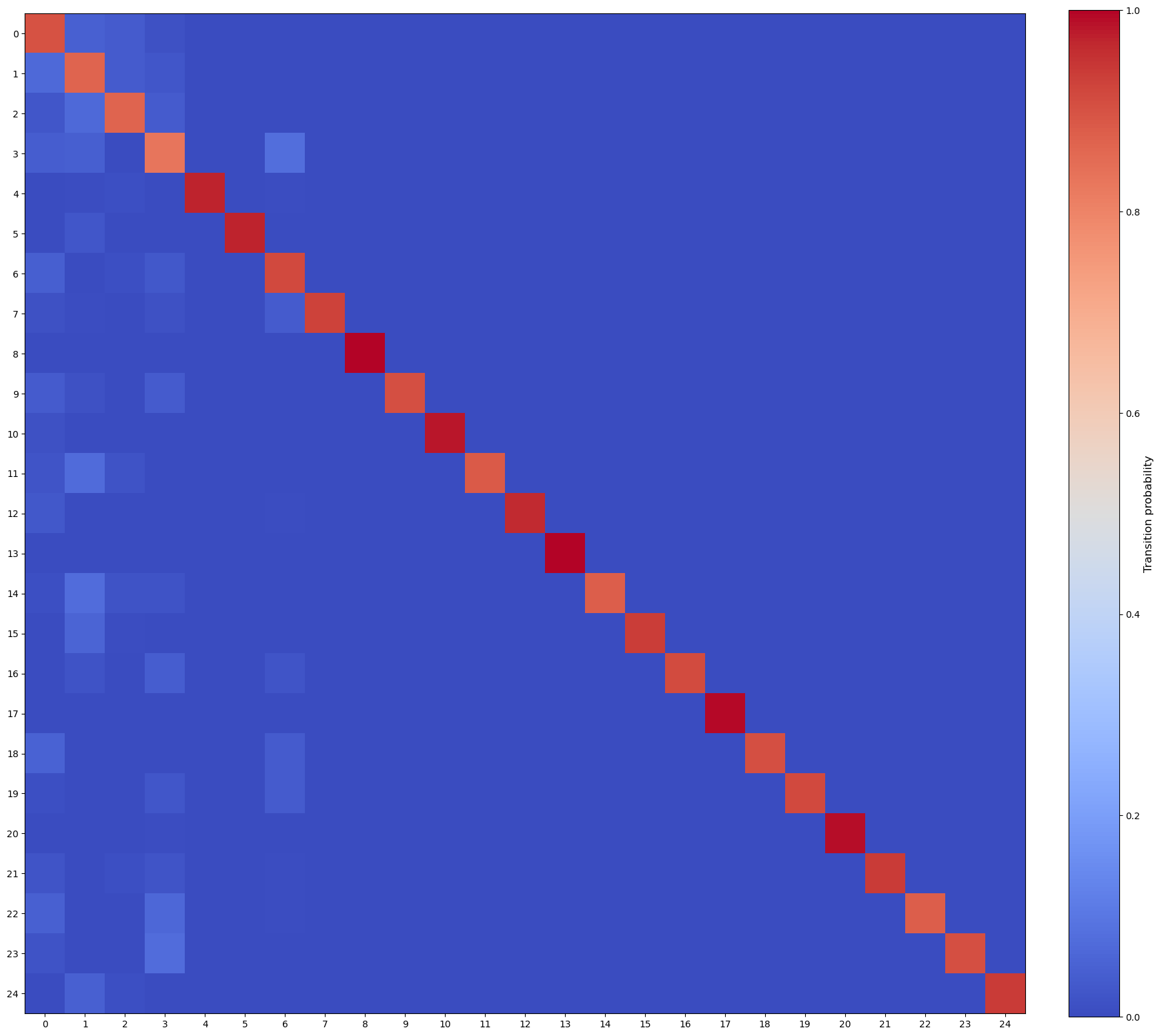}
    \caption{Transition probabilities among the 25 most active relations in the $R=25$ relation-cardinality setting of ICEWS18.}
    \label{fig:pi_ICEWS}
\end{figure}

We conduct a mechanism-oriented relation-cardinality ablation on
ICEWS18 to examine how relation coupling behaves as the number of
retained relation types increases. D-NGPPF and RENet are selected as the strongest probabilistic and neural baselines, respectively, in the main comparison.
Fig.~\ref{fig:auc_comparison} reports the performance under varying
numbers of relations $R$. The AUC-PR of PGRE initially increases,
reaches its maximum at $R=25$, and then gradually decreases, while
remaining consistently higher than those of the competing methods.
Its AUC-ROC exhibits an overall upward trend, with a slight fluctuation
at $R=75$. In contrast, the AUC-PR values of the competing methods
generally decline as $R$ increases, suggesting that they benefit less
from the additional cross-relation structure.

The ablation variant PGE, which removes the relation transition matrix,
exhibits a substantial reduction in AUC-PR across all settings,
demonstrating the contribution of explicit relation coupling.
Similarly, the AUC-PR values of D-NGPPF and RENet decrease as $R$
grows, whereas their AUC-ROC improvements are less pronounced than
those of PGRE. These results indicate that the proposed
relation-transition mechanism helps PGRE maintain more robust
predictive performance as relation diversity increases.

To further examine the learned structure, we visualize the temporal evolution
of latent factors and the relation transition matrix $\Pi_{r,r'}$.
Fig.~\ref{fig:rk} illustrates the temporal evolution of latent factors for
representative positive interaction types. Certain latent factors remain
consistently active across multiple relations, suggesting the presence of
shared latent communities that drive their temporal dynamics.
Fig.~\ref{fig:pi_ICEWS} shows the transition probabilities in the $R=25$ setting. The matrix displays a dominant diagonal pattern,
reflecting strong temporal self-dependence of relations across adjacent time
steps. In addition, several off-diagonal bands reveal structured transitions
between semantically related relations, indicating dynamic coupling beyond
self-persistence. These observations demonstrate that PGRE captures both
interpretable latent evolution and structured inter-relational interactions,
which jointly contribute to its robustness in sparse multi-relational
settings.

\section{Conclusion}

We have proposed a probabilistic network model for capturing temporal evolution and relational interactions in dynamic knowledge graphs.
In contrast to existing Bayesian network models, the proposed framework is directly applicable to dynamic knowledge graph completion.
The model has introduced a structured relation transition kernel to characterize inter-relational dynamics within latent communities and has supported efficient closed-form posterior inference via Gibbs sampling.
Experimental results on multiple temporal knowledge graph benchmarks have demonstrated competitive relation prediction performance, with
substantial AUC-PR improvements on ICEWS18 and GDELT, together with interpretable relational and temporal structures revealed through ablation and visualization analyses.

Future work includes extending the model to settings with rapidly evolving event streams and learning time-varying relation transition dynamics.
Another promising direction is adapting the framework to heterogeneous temporal networks with irregular or partially overlapping node sets.



\begin{acknowledgements} 

%

This work was partially supported by the National Natural Science Foundation of China (NSFC) (Grant Nos. 62476047 and 62276171), the Peking University Mathematics Challenge Funding Program (Grant No. 2024SRMC10), the Shenzhen Science and Technology Program (Grant Nos. ZDCY20250901110940006 and JCYJ20240813141503005), the Dongguan Key Laboratory for AI and Dynamical Systems, the Dongguan Key Laboratory for Intelligence and Information Technology, the Dongguan Key Laboratory for Data Science and Intelligent Medicine, the Guangdong Research Team for Communication and Sensing Integrated with Intelligent Computing (Project No. 2024KCXTD047), and the Guangdong Multidisciplinary Innovation Research Group for New-Generation Intelligent Systems and Diagnostic-Therapeutic Applications (Grant No. 2025KCXTD031).

\end{acknowledgements}

\bibliography{uai2026-template}

\newpage

\onecolumn

\title{Poisson–Gamma Modeling of Inter-Relational Dependencies in Dynamic Knowledge Graphs\\(Supplementary Material)}
\makeatletter
\let\@thanks\@empty
\makeatother

\maketitle

\section{Supplementary Material}

\subsection{Baseline Implementations}

We provide additional details on the baseline models used in the experiments.

\paragraph{Probabilistic Models for Dynamic Networks.}
DPGM, D-NGPPF, and G-HSEPM are designed for dynamic single-relation networks.
Following standard practice, we train one independent model for each relation.
For directed relations, each relational graph is converted into an undirected bipartite graph by separating entities into source and target roles, resulting in a $(2N \times 2N)$ adjacency matrix per relation.

\paragraph{Probabilistic Models for Dynamic Count Data.}
PGDS, PRGDS, and NBRGDS operate on temporal count tensors.
We reshape the original $R \times T \times N \times N$ temporal knowledge graph into an event--time matrix, where each event corresponds to a triple $(s, r, o)$.
Model inference is conducted following the settings described in the original papers.

\paragraph{Neural Models for Temporal Knowledge Graphs.}
Know-Evolve, DyRep, and RENet are neural temporal knowledge graph models that output real-valued scores for candidate events.
We use the official implementations when available and follow the default hyperparameter configurations recommended by the authors.
For models that do not directly produce probabilities, a sigmoid function is applied to convert scores into probabilistic predictions.

\subsection{Additional Experimental Settings}
All probabilistic generative models, including PGRE and count-based baselines, are trained using Gibbs sampling.
For all datasets, the number of latent communities is fixed to $K=200$.
To ensure fair comparison, the remaining hyperparameters for each probabilistic model follow the recommendations in the original papers or official implementations.
Model inference is performed with 2000 Gibbs sampling iterations, where the first 1000 iterations are discarded as burn-in and the remaining 1000 iterations are used for posterior estimation.

For neural network-based methods, we use the authors’ publicly released code and follow their recommended experimental settings.
All neural models are trained in a supervised learning setting using the binary cross-entropy loss.
Training is conducted for up to 1000 epochs with early stopping based on validation performance, using a patience of 20.
The model checkpoint achieving the best validation performance is selected for testing.
Unless otherwise specified, the learning rate and batch size are set to 0.0001 and 200, respectively.

To reduce the effect of randomness, each experiment was repeated five times with different random seeds, and the reported results correspond to average performance.
All experiments are implemented in Python 3.9.12 with PyTorch 2.1.0 and CUDA 12.1, and are conducted on a Dell Precision 7920 workstation running Ubuntu Linux (kernel 6.11.0-25-generic).
The implementation code and experimental scripts are publicly available at \url{https://github.com/ffffkgh/PGRE}.

\subsection{Sampling Diagnostics}
\label{sec:convergence}

We assessed the sampling behavior of the Gibbs sampler using effective
sample sizes (ESSs) and trace plots for a subset of representative model
parameters. The sampler was run for 2,000 iterations, with the first 1,000
iterations discarded as burn-in and the remaining 1,000 iterations retained
for posterior estimation. Because the model contains a large number of latent
variables, we report diagnostics for selected entries of the relation
transition matrix and selected normalized temporal community weights on a
representative dataset and temporal--relation slice.

For the relation transition parameters, the post-burn-in ESS values were
49.97 for $\pi_{00}$, 336.70 for $\pi_{20}$, and 246.59 for
$\pi_{70}$. We additionally monitored two representative normalized temporal
community weights,
\[
\widetilde{\delta}_{k}^{(t,r)}
=
\frac{\delta_k^{(t,r)}}
{\sum_{k'=1}^{K}\delta_{k'}^{(t,r)}},
\]
whose post-burn-in ESS values were 23.63 for
$\widetilde{\delta}_{61}^{(4,0)}$ and 12.99 for
$\widetilde{\delta}_{24}^{(4,0)}$. The corresponding results are summarized
in Table~\ref{tab:ess}.

\begin{table}
    \centering
    \caption{Post-burn-in effective sample sizes for selected representative
    model parameters.}
    \label{tab:ess}
    \begin{tabular}{lc}
        \toprule
        Parameter & ESS \\
        \midrule
        $\pi_{00}$ & 49.97 \\
        $\pi_{20}$ & 336.70 \\
        $\pi_{70}$ & 246.59 \\
        $\widetilde{\delta}_{61}^{(4,0)}$ & 23.63 \\
        $\widetilde{\delta}_{24}^{(4,0)}$ & 12.99 \\
        \bottomrule
    \end{tabular}
\end{table}

\begin{figure*}
    \centering
    \includegraphics[width=0.75\textwidth]{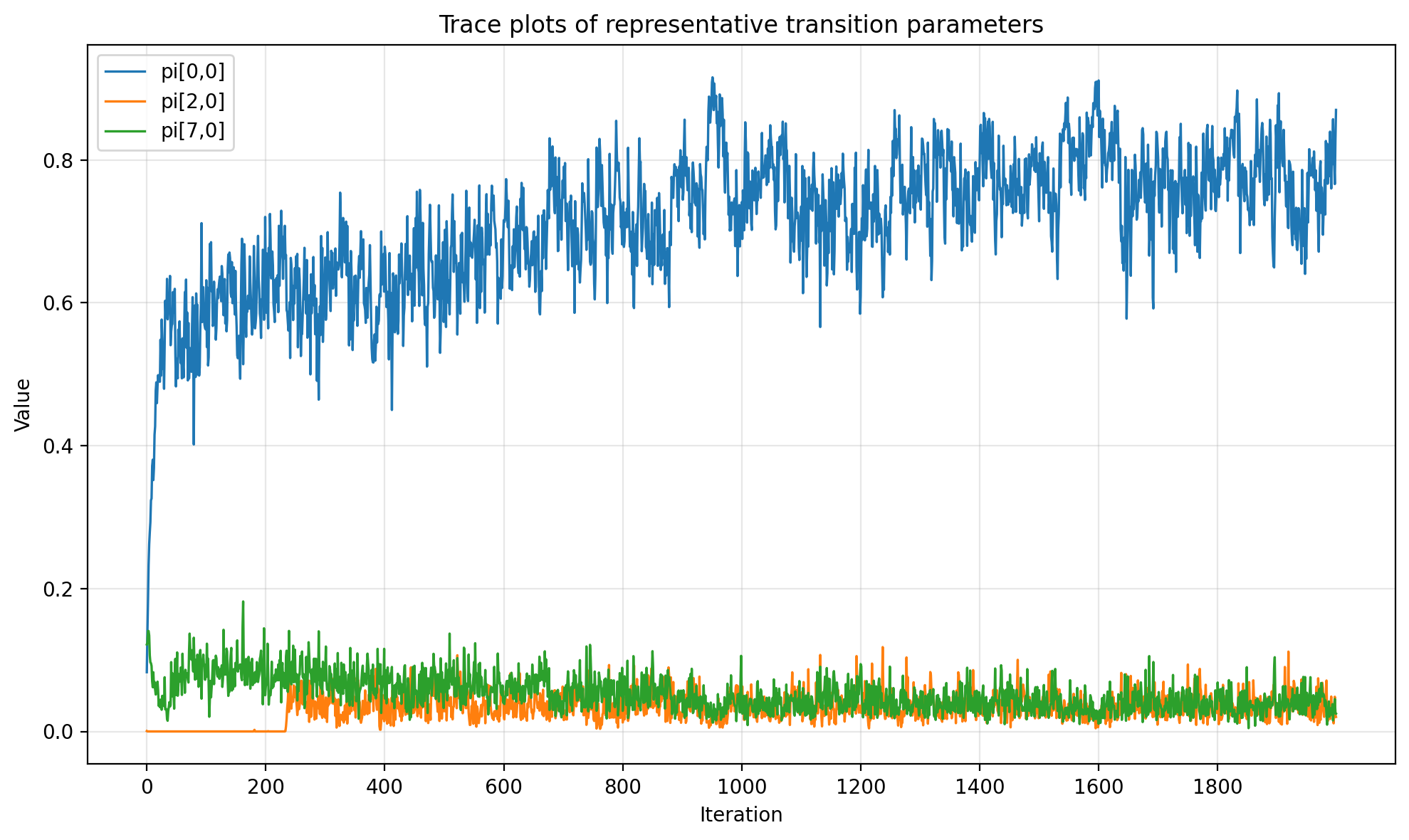}
    \caption{Trace plots of selected representative relation transition
    parameters over 2,000 Gibbs sampling iterations. The first 1,000
    iterations are treated as burn-in, and the remaining 1,000 iterations
    are retained for posterior estimation. The trajectories exhibit an
    initial adaptation period followed by comparatively more stable
    post-burn-in behavior.}
    \label{fig:trace_pi}
\end{figure*}

\begin{figure*}
    \centering
    \includegraphics[width=0.75\textwidth]{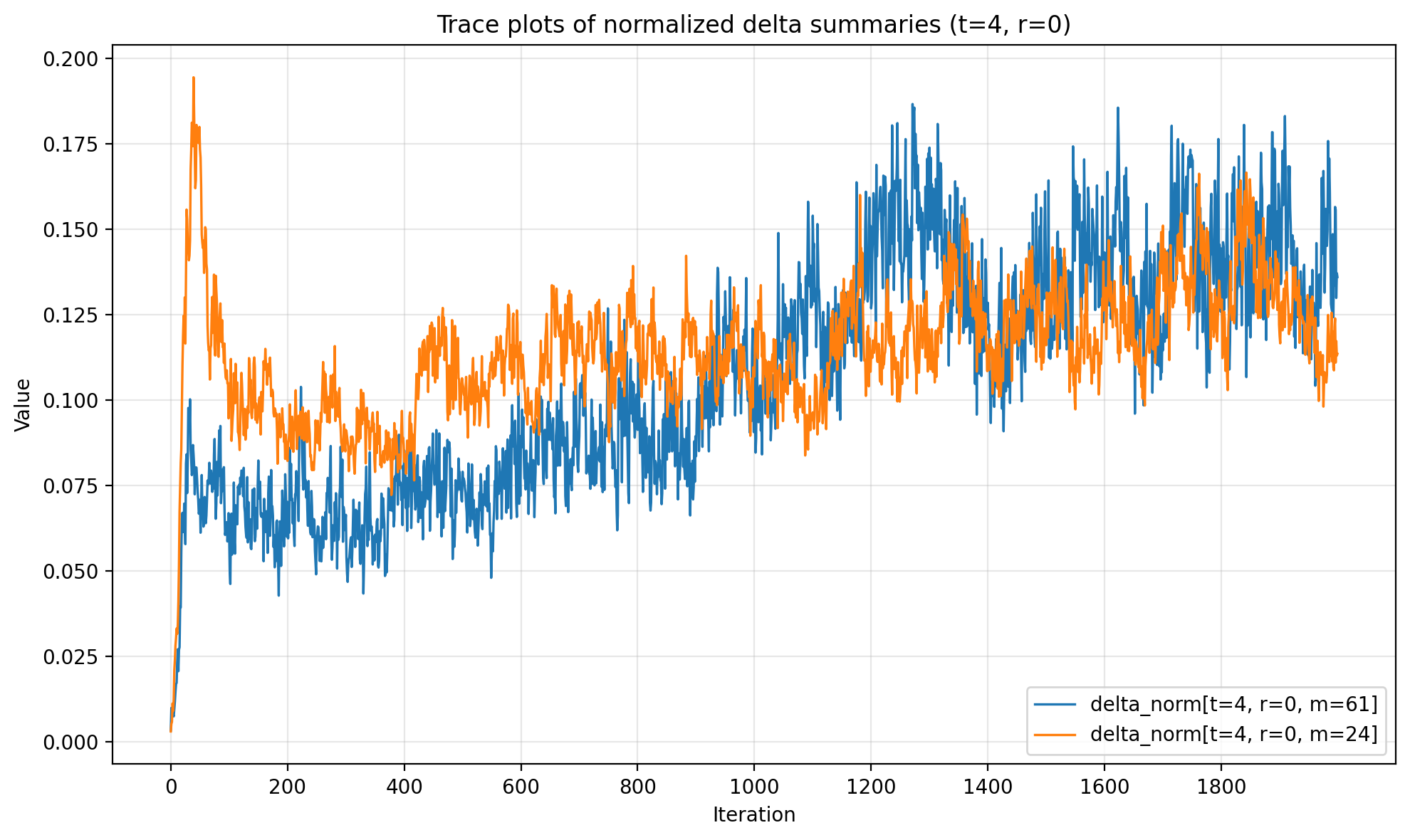}
    \caption{Trace plots of selected normalized temporal community weights
    at $(t,r)=(4,0)$ over 2,000 Gibbs sampling iterations. Compared with the
    transition parameters, these summaries exhibit stronger serial dependence
    and slower variation, consistent with their lower post-burn-in ESS
    values.}
    \label{fig:trace_delta}
\end{figure*}

Figs.~\ref{fig:trace_pi} and~\ref{fig:trace_delta} show the corresponding
trace plots over all 2,000 iterations. The representative transition
parameters exhibit an initial adaptation period followed by comparatively
more stable trajectories in the retained sampling period. Their ESS values
also indicate more effective mixing for $\pi_{20}$ and $\pi_{70}$, while
$\pi_{00}$ exhibits stronger serial dependence.

The normalized temporal community-weight summaries display slower variation
and stronger autocorrelation, consistent with their lower ESS values.
Nevertheless, their trajectories become comparatively more stable after the
initial adaptation period. Taken together, the trace plots and ESS values
provide complementary evidence that the adopted sampling design captures a
more stable posterior regime after burn-in. These diagnostics therefore
support the use of 1,000 burn-in iterations followed by 1,000 retained
iterations for posterior estimation, while also indicating that some latent
community-weight summaries remain more strongly autocorrelated than the
transition parameters.

\subsection{Sensitivity Analysis of the Temporal Scale Parameter}
\label{sec:tau_sensitivity}

We examined the sensitivity of PGRE to the temporal scale parameter
$\tau$, which controls the scale of the Gamma--Markov evolution. We varied
$\tau$ over $\{0.5, 1.0, 1.5, 2.0\}$ while keeping all other experimental
settings unchanged. The results are reported in
Table~\ref{tab:tau_sensitivity}.

\begin{table}
    \centering
    \caption{Sensitivity of PGRE to the temporal scale parameter $\tau$.
    The results correspond to a single-run sensitivity analysis.}
    \label{tab:tau_sensitivity}
    \begin{tabular}{ccc}
        \toprule
        $\tau$ & AUC-ROC & AUC-PR \\
        \midrule
        0.5 & 0.9324 & \textbf{0.2067} \\
        1.0 & 0.9327 & 0.1991 \\
        1.5 & \textbf{0.9329} & 0.2002 \\
        2.0 & 0.9316 & 0.2008 \\
        \bottomrule
    \end{tabular}
\end{table}

The performance remains stable across the tested values of $\tau$.
Specifically, the AUC-ROC values vary only from 0.9316 to 0.9329, while
the AUC-PR values range from 0.1991 to 0.2067. Although $\tau=0.5$
achieves the highest AUC-PR and $\tau=1.5$ gives the highest AUC-ROC,
the differences are modest. These results indicate that PGRE is reasonably
robust to moderate variations in the temporal scale parameter and does not
rely on a narrowly tuned value of $\tau$. We therefore use $\tau=1.0$ as
the default setting in the main experiments.

\subsection{Markov Chain Monte Carlo Inference}
Here we present the Gibbs sampling for the PGRE model. 

\noindent \textbf{Sampling latent counts $x_{ij}^{(t,r)}$.} We sample the latent count associated with each observed event $\mathcal{M}_{ij}^{(t,r)}$ as

\begin{equation}
    (x_{ij}^{(t,r)}|-) \sim \mathcal{M}_{ij}^{(t,r)}\text{Po}_{+}(\sum_{k=1}^{K}\delta_{k}^{(t,r)} \phi_{ik} \psi_{jk}).
    \label{xij}
\end{equation}
where $\mathrm{Pois}_{+}(\cdot)$ denotes a zero-truncated Poisson distribution. 
Since $x_{ij}^{(t,r)} = \sum_{k=1}^{K} x_{ijk}^{(t,r)}$, by the additive property of the Poisson distribution we have $x_{ijk}^{(t,r)} \sim \text{Pois} (\delta_{k}^{(t,r)} \phi_{ik} \psi_{jk})$. According to the Poisson--multinomial equivalence (Theorem~\ref{thm:pm}), the latent count $x_{ijk}^{(t,r)}$ can be sampled as

\begin{equation}
    (x_{ijk}^{(t,r)}|-) \sim \text{Mult} (x_{ij}^{(t,r)},(\frac{\delta_{k}^{(t,r)} \phi_{ik} \psi_{jk}}{\sum_{k=1}^K \delta_{k}^{(t,r)} \phi_{ik} \psi_{jk}})).
    \label{xijk}
\end{equation}

\noindent \textbf{Sampling community factor loadings $\phi_{ik}$ and $\psi_{jk}$.} By the Gamma--Poisson conjugacy, the posterior distributions of the subject- and object-side memberships are given by

\begin{equation}
\begin{split}
(\phi_{ik}\mid -) \sim \mathrm{Gam}\Biggl(
    & a_0 + 
    \sum_{t=1}^{T} \sum_{r=1}^{R} 
    \sum_{\substack{j=1 \\ j \neq i}}^{N} 
    x_{ijk}^{(t,r)}, \\
    & 1/\Bigl(
        c_i + 
        \sum_{t=1}^{T} \sum_{r=1}^{R} 
        \sum_{\substack{j=1 \\ j \neq i}}^{N} 
        \delta_{k}^{(t,r)} \psi_{jk} 
    \Bigr)
\Biggr),
\end{split}
\label{phi}
\end{equation}

\begin{equation}
\begin{split}
(\psi_{jk} \mid -) \sim \mathrm{Gam}\Biggl(
    & a_1 + 
    \sum_{t=1}^{T} \sum_{r=1}^{R} 
    \sum_{\substack{i=1 \\ i \neq j}}^{N} 
    x_{ijk}^{(t,r)}, \\
    & 1/\Bigl(
        c_j + 
        \sum_{t=1}^{T} \sum_{r=1}^{R} 
        \sum_{\substack{i=1 \\ i \neq j}}^{N} 
        \delta_{k}^{(t,r)} \phi_{ik}
    \Bigr)
\Biggr).
\end{split}
\label{psi}
\end{equation}

\noindent \textbf{Sampling entity-specific scaling parameters $c_i$ and $c_j$.} 
We place Gamma priors over the entity-specific scaling parameters as 
$c_i \sim \mathrm{Gam}(f_0, 1/g_0)$ and $c_j \sim \mathrm{Gam}(z_0, 1/k_0)$. 
Using the Gamma--Gamma conjugacy, the posterior distributions can be derived in closed form as:

\begin{equation}
    (c_i|-) \sim \text{Gam} (f_0 + Ka_0, 1/(g_0 + \sum_{k=1}^{K}\phi_{ik})),
    \label{eq:ci}
\end{equation}
\begin{equation}
    (c_j|-) \sim \text{Gam} (z_0 + K a_1, 1/(k_0 + \sum_{k=1}^{K}\psi_{jk})),
    \label{eq:cj}
\end{equation}
where $K$ denotes the number of latent communities.

\noindent \textbf{Sampling community weights $\delta_{k}^{(t,r)}$.} 
The community weights $\delta_{k}^{(t,r)}$ evolve under a Markovian structure, so both backward and forward information must be incorporated into the posterior updates.

For the last time step $t = T$, we define the aggregated latent count as $x_{\cdot \cdot k}^{(T,r)} = \sum_{i=1}^{N} \sum_{\substack{j=1 \\ j \neq i}}^{N} x_{ijk}^{(T,r)}$, which follows a Poisson distribution $x_{\cdot \cdot k}^{(T,r)} \sim \text{Pois} (\delta_{k}^{(T,r)} s_k)$, where $s_k = \sum_{i=1}^{N} \sum_{\substack{j=1 \\ j \neq i}}^{N} \phi_{ik}\psi_{jk}$.  
By applying the gamma–Poisson conjugacy, the posterior of $\delta_{k}^{(T,r)}$ is given by

\begin{equation}
    (\delta_{k}^{(T,r)}|-) \sim \text{Gam} \left( x_{\cdot \cdot k}^{(T,r)} + \sum_{r'=1}^{R} \pi_{rr'} \delta_{k}^{(T-1,r')}, 1/(\tau + s_k) \right).
    \label{delta_kT}
\end{equation}

For $t = T - 1$, to incorporate forward information, we marginalize out $\delta_{k}^{(T,r)}$ using Theorem~\ref{thm:gp}, which leads to

\begin{equation}
    x_{\cdot \cdot k}^{(T,r)} \sim \text{NB} \left( \sum_{r'=1}^{R} \pi_{rr'} \delta_{k}^{(T-1,r')}, s_k / (\tau + s_k) \right).
\end{equation}

According to Theorem~\ref{thm:pl}, the negative binomial distribution can be augmented with an auxiliary variable as

\begin{equation}
    l_{k}^{(T,r)} \sim \text{CRT} \left( x_{\cdot \cdot k}^{(T,r)}, \sum_{r'=1}^{R} \pi_{rr'} \delta_{k}^{(T-1,r')} \right).
    \label{lTk}
\end{equation}

The joint distribution of $x_{\cdot \cdot k}^{(T,r)}$ and $l_{k}^{(T,r)}$ can then be written as
\begin{equation}
    x_{\cdot \cdot k}^{(T,r)} \sim \text{SumLog}(l_{k}^{(T,r)}, \rho_k^{(T,r)}),
\end{equation}

\begin{equation}
    l_{k}^{(T,r)} \sim \text{Pois}\left( -\sum_{r'=1}^{R} \pi_{rr'} \delta_{k}^{(T-1,r')} \ln(1 - \rho_k^{(T)}) \right),
\end{equation}
where $\rho_k^{(T)} = \frac{s_k}{\tau + s_k}$. Since $l_k^{(T,r)} = l_{k}^{(T,r\cdot)} = \sum_{r_2=1}^{R} l_{k}^{(T,rr_2)}$, the distribution of $l_{k}^{(T,rr')}$ is

\begin{equation}
l_{k}^{(T,rr')} \sim \text{Pois}\left( -\pi_{rr'} \delta_k^{(T-1,r')} \ln(1 - \rho_k^{(T)}) \right).
\end{equation}

Using the additive property of the Poisson distribution and $\sum_{r_1=1}^{R}\pi_{r_1r}=1$, we have 
\begin{equation}
l_{k}^{(T,\cdot r)} \sim \text{Pois}\left( -\delta_k^{(T-1,r)} \ln(1 - \rho_k^{(T)}) \right).
\end{equation}

Given $x_{\cdot \cdot k}^{(T-1,r)} \sim \text{Pois} (\delta_k^{(T-1,r)} s_k)$, the Poisson additive property leads to
\begin{equation}
    x_{\cdot \cdot k}^{(T-1,r)} + l_{k}^{(T,\cdot r)} \sim \text{Pois} \left( \delta_k^{(T-1,r)} ( s_k - \ln(1 - \rho_k^{(T)}) ) \right).
\end{equation}

This combines backward and forward information at $t = T - 1$. With the gamma prior on $\delta_k^{(T-1,r)}$, its conditional posterior is

\begin{equation}
\begin{aligned}
    (\delta_k^{(T-1,r)}|-) &\sim \text{Gam} \left( x_{\cdot \cdot k}^{(T-1,r)} + l_{k}^{(T,\cdot r)} + \sum_{r'}^{R}\pi_{rr'}\delta_{k}^{(T-2,r')}, \right. \\
    &\quad \left. 1 / (\tau + s_k - \ln(1 - \rho_k^{(T)})) \right).
\end{aligned}
\end{equation}

For intermediate steps $t = T - 2, \dots, 2$, we introduce
\begin{equation}
    l_{k}^{(t,\cdot r)} \sim \text{CRT} \left( x_{\cdot \cdot k}^{(t,r)} + l_{k}^{(t+1,\cdot r)}, \sum_{r'=1}^{R} \pi_{rr'} \delta_{k}^{(t-1,r')} \right),
    \label{ltk}
\end{equation}
and apply the same augmentation–sampling procedure:

\begin{equation}
\begin{aligned}
    (\delta_k^{(t,r)}|-) &\sim \text{Gam} \left( x_{\cdot \cdot k}^{(t,r)} + l_{k}^{(t+1,\cdot r)} + \sum_{r'}^{R}\pi_{rr'}\delta_{k}^{(t-1,r')}, \right. \\
    &\quad \left. 1 / (\tau + s_k - \ln(1 - \rho_k^{(t+1)})) \right),
    \label{delta_kt}
\end{aligned}
\end{equation}
where 
\begin{equation}
    \rho_k^{(t+1)} = \frac{s_k - \ln(1 - \rho_k^{(t+2)})}{\tau + s_k - \ln(1 - \rho_k^{(t+2)})}. 
    \label{rho}
\end{equation}

Finally, for $t = 1$, we augment

\begin{equation}
    l_{k}^{(1,r)} \sim \text{CRT} \left( x_{\cdot \cdot k}^{(1,r)} + l_{k}^{(2,\cdot r)}, \nu_r / K \right),
    \label{l1k}
\end{equation}
and sample

\begin{equation}
\begin{aligned}
    (\delta_k^{(1,r)}|-) &\sim \text{Gam} \left( x_{\cdot \cdot k}^{(1,r)} + l_{k}^{(2,\cdot r)} + \nu_r / K, \right. \\
    &\quad \left. 1 / (\tau + s_k - \ln(1 - \rho_k^{(2)})) \right).
    \label{delta_k1}
\end{aligned}
\end{equation}

\noindent \textbf{Sampling the transition matrix $\pi_r$.}  
After marginalizing out $\delta$, the auxiliary counts follow a multinomial distribution $(l_{k}^{(t,1r)}, \dots, l_{k}^{(t,Rr)}) \sim \text{Mult}\left(l_{k}^{(t,\cdot r)}, (\pi_{1r}, \dots, \pi_{Rr})\right)$. By the Dirichlet–multinomial conjugacy, the posterior distribution of $\pi_r$ is given by

\begin{equation}
\begin{aligned}
    (\pi_r|-) \sim \text{Dir}\Bigl(
        & \nu_1 \nu_r + \sum_{t=2}^{T} \sum_{k=1}^{K} l_{k}^{(t,1r)}, \ \dots, \\
        & \xi \nu_r + \sum_{t=2}^{T} \sum_{k=1}^{K} l_{k}^{(t,rr)}, \ \dots, \\
        & \nu_R \nu_r + \sum_{t=2}^{T} \sum_{k=1}^{K} l_{k}^{(t,Rr)}
    \Bigr).
    \label{pi}
\end{aligned}
\end{equation}

\noindent \textbf{Sampling $\nu_r$ and $\xi$.}  
We marginalize over $\Pi$ to obtain a Dirichlet–multinomial distribution

\begin{equation}
\begin{aligned}
    & ( l_{\cdot}^{(\cdot,1r)}, \dots, l_{\cdot}^{(\cdot,Rr)} ) \\
    &\quad \sim \text{DirMult}\Bigl(
        l_{\cdot}^{(\cdot,\cdot r)}, \ 
        ( \nu_1\nu_r, \dots, \xi\nu_r, \dots, \nu_R\nu_r )
    \Bigr),
\end{aligned}
\end{equation}
where $l_{\cdot}^{(\cdot,\cdot r)} = \sum_{t=1}^{T} \sum_{k=1}^{K} \sum_{r_1=1}^{K}l_{k}^{(t,r_1 r)}$. 

By introducing a beta-distributed auxiliary variable $q_r$, the Dirichlet–multinomial distribution can be rewritten as a negative binomial distribution:
\begin{equation}
    q_r \sim \text{Beta}\left( l_{\cdot}^{(\cdot,\cdot r)}, \, \nu_r \left( \xi + \sum_{r' \neq r} \nu_{r'} \right) \right).
    \label{qr}
\end{equation}

Then, $l_{\cdot}^{(\cdot,rr)}$ and $l_{\cdot}^{(\cdot, rr')}$ follow negative binomial distributions as 
$l_{\cdot}^{(\cdot,rr)} \sim \mathrm{NB}(\xi\nu_r, q_r)$ and 
$l_{\cdot}^{(\cdot, rr')} \sim \mathrm{NB}(\nu_{r}\nu_{r'}, q_r)$.
To further facilitate posterior inference, we introduce auxiliary count variables $h_{rr}$ and $h_{rr'}$ via the Chinese Restaurant Table (CRT) distribution:
\begin{equation}
\begin{aligned}
    h_{rr} &\sim \text{CRT}\left( l_{\cdot}^{(\cdot,rr)}, \, \xi \nu_r \right), \\
    h_{rr'} &\sim \text{CRT}\left( l_{\cdot}^{(\cdot,rr')}, \, \nu_{r}\nu_{r'} \right).
\end{aligned}
\label{hrr2}
\end{equation}

Using the gamma–Poisson conjugacy, $\xi$ is then sampled as
\begin{equation}
    (\xi|-) \sim \mathrm{Gam}\left( b_0 + \sum_{r=1}^{R} h_{rr}, \, \frac{1}{ e_0 - \sum_{r=1}^{R} \nu_r \ln (1 - q_r) } \right).
    \label{xi2}
\end{equation}

Next, we define 
\[
    n_r = h_{rr} + \sum_{r_1 \neq r}h_{r_{1}r} + \sum_{r_2 \neq r}h_{rr_{2}} + \sum_{k=1}^{K}l_{k}^{(1,r \cdot)},
\]
where $l_{k}^{(1,r)} \sim \text{Pois} (\tau\nu_r\ln{(1-\rho_k^{(1)})})$. Using the Poisson additive property and gamma–Poisson conjugacy, we have

\begin{equation}
    (\nu_r|-) \sim \text{Gam} (\frac{\gamma_0}{R} + n_r, 1/(\beta + t_r)),
    \label{nu2}
\end{equation}
where $t_r = -\ln(1 - q_r)\left(\xi + \sum_{r_{1} \neq r} \nu_{r_{1}}\right) 
- \sum_{r_{2} \neq r} \ln(1 - q_{r_{2}}) \nu_{r_{2}} 
- \sum_{k=1}^{K} \frac{1}{K} \ln(1 - \rho_k^{(1)})$.

\noindent \textbf{Sampling $\beta$.}  
Finally, by gamma–gamma conjugacy, the posterior of $\beta$ is derived as

\begin{equation}
    (\beta|-) \sim \text{Gam} (d_0+\gamma_0, 1/(h_0 + \sum_{r=1}^{R}\nu_r)).
    \label{beta2}
\end{equation}

Algorithm \ref{algorithm} summarises the full sampling procedure of PGRE.
\begin{algorithm}[!ht]
    \caption{Gibbs sampling algorithm for PGRE}
    \label{algorithm}
    \begin{algorithmic}[1]
        \Require Dynamic relational data $G^{(1)}, \cdots, G^{(T)}$.
        \State Initialise the number of communities $K$ and other parameters;
        \Repeat
            \State Sample $x_{ij}^{(t,r)}$ (Eq.~\ref{xij}) and update $x_{ijk}^{(t,r)}$ (Eq.~\ref{xijk})
            \State Sample $\phi_{ik}$ (Eq.~\ref{phi}) and $\psi_{jk}$ (Eq.~\ref{psi})
            \State update $c_{i}$ (Eq.~\ref{eq:ci}) and $c_{j}$ (Eq.~\ref{eq:cj})
            
            \For{$t=T,\cdots,1$}
                \If{$t = T$}
                    \State Sample $l_{k}^{(T,r)}$ (Eq.~\ref{lTk})
                \ElsIf{$t = T-1, \cdots, 2$}
                    \State Sample $l_{k}^{(t,r)}$ (Eq.~\ref{ltk})
                \ElsIf{$t = 1$}
                    \State Sample $l_{k}^{(1,r)}$ (Eq.~\ref{l1k})
                \EndIf
                \State Update $\rho_k^{(t)}$ (Eq.~\ref{rho})
            \EndFor
            
            \For{$t=1,\cdots,T$}
                \If{$t = 1$}
                    \State Sample $\delta_k^{(1,r)}$ (Eq.~\ref{delta_k1})
                \ElsIf{$t = 2, \cdots, T-1$}
                    \State Sample $\delta_k^{(t,r)}$ (Eq.~\ref{delta_kt})
                \ElsIf{$t = T$}
                    \State Sample $\delta_{k}^{(T,r)}$ (Eq.~\ref{delta_kT})
                \EndIf
            \EndFor

            \State Sample $\pi_r$ (Eq.~\ref{pi}), update $q_r$ (Eq.~\ref{qr}) and $h_{rr}$ (Eq.~\ref{hrr2})
            \State Sample $\xi$ (Eq.~\ref{xi2}) and $\nu_{r}$ (Eq.~\ref{nu2})
            \State Sample $\beta$ (Eq.~\ref{beta2})
            
        \Until{convergence}
        \Ensure Posterior mean of $\phi_{ik}$, $\psi_{jk}$ and $\delta_k^{(t,r)}$
    \end{algorithmic}
\end{algorithm}

\subsection{Full Probabilistic Graphical Model}
\begin{figure*}
    \centering
    \includegraphics[width=0.8\textwidth]{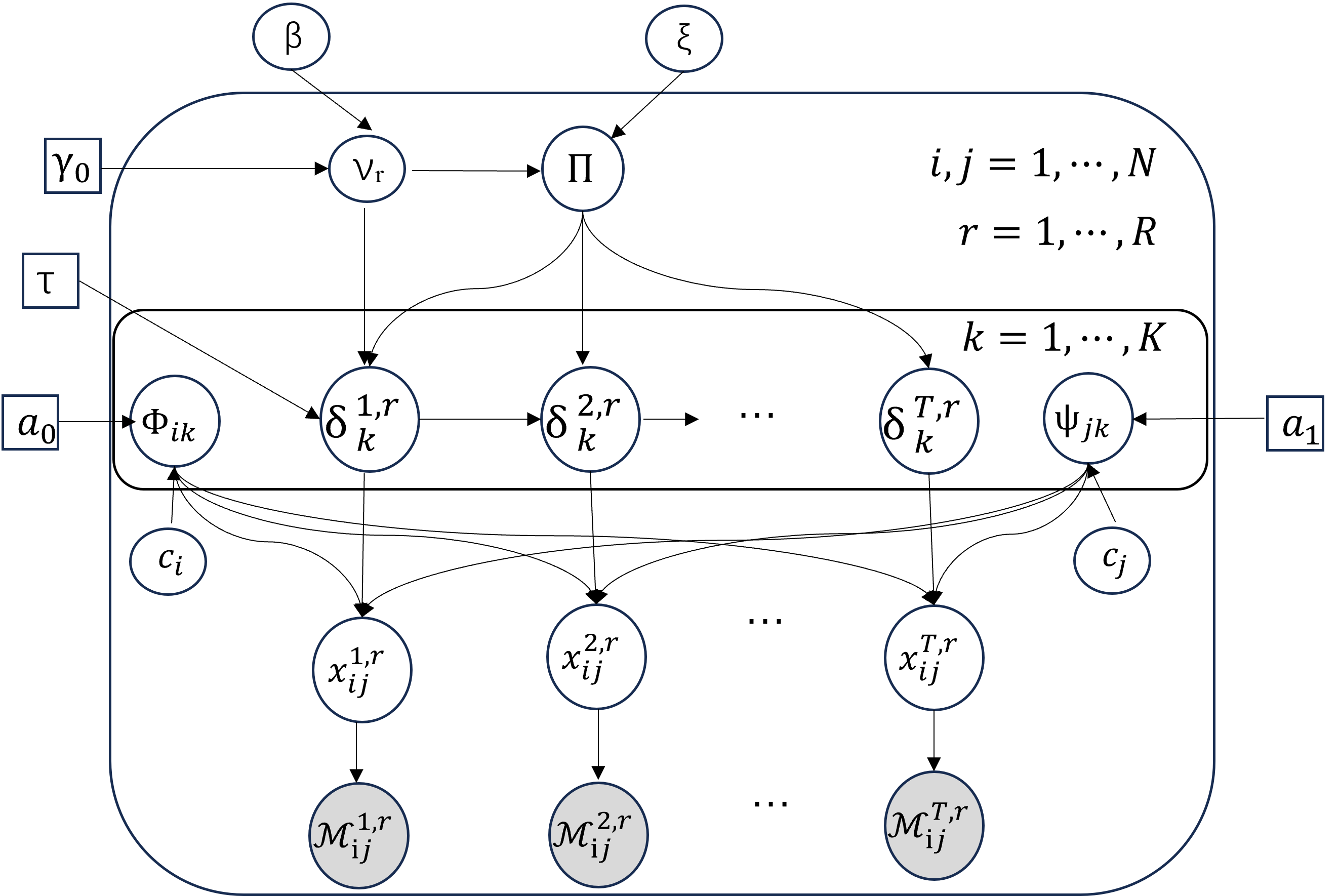}
    \caption{Full probabilistic graphical model of PGRE.}
    \label{fig:full_pgm}
\end{figure*}

Fig.~\ref{fig:full_pgm} presents the full probabilistic graphical model of PGRE,
which illustrates the complete generative process including entity-level
latent factors, relation-level temporal evolution, and observed interactions.

\appendix

\end{document}